\documentclass[lettersize,journal]{IEEEtran}%

\usepackage{xcolor}

\usepackage{amsfonts}
\usepackage{amsmath}
\usepackage{amsthm}
\usepackage{mathtools}
\usepackage{hyperref}
\usepackage[ruled,vlined,linesnumbered]{algorithm2e}

\usepackage{cleveref}
\crefname{subequation}{Eq.}{Eqs.}

\newcommand{\defeq}{\vcentcolon=}
\newcommand{\matr}[1]{\mathbf{#1}}
\newtheorem{dfn}{Definition}
\newtheorem{lemma}{Lemma}

\usepackage[font=small,labelfont=bf]{caption}
\usepackage{cite}
\usepackage[normalem]{ulem}
\usepackage{comment}
\usepackage{enumitem}
\DeclarePairedDelimiter{\norm}{\lVert}{\rVert}

\usepackage{lipsum}
\usepackage{float}
\usepackage[font=small, labelfont=bf]{subfig}
\usepackage{svg}
\usepackage{tikz}
\usepackage{makecell}
\DeclareMathOperator*{\argmin}{arg\,min}
\newtheorem*{lemma*}{Lemma}

\hypersetup{
    colorlinks=true,
    linkcolor=blue,
    filecolor=green,      
    urlcolor=magenta,
    citecolor=teal,
}

\begin{document}
%
\title{Approximately Optimal Global Planning \\
for Contact-Rich SE(2) Manipulation \\
on a Graph of Reachable Sets\thanks{This manuscript is under submission to IEEE Transactions on Robotics.}}
%
%
%
\author{Simin Liu$^{1,2,*}$, Tong Zhao$^{2, *}$, Bernhard Paus Graesdal$^{3}$, Peter Werner$^{3}$, Jiuguang Wang$^{2}$,  John Dolan$^{1}$, Changliu Liu$^{1}$, Tao Pang$^{2}$ \thanks{ $^{1}$Robotics Institute, Carnegie Mellon University \\ 
\indent $^{2}$Robotics and AI Institute\\
\indent $^{3}$CSAIL, Massachusetts Institute of Technology
}}

\maketitle

\begin{abstract}
If we consider human manipulation, it is clear that contact-rich manipulation (CRM)-the ability to use any surface of the manipulator to make contact with objects-can be far more efficient and natural than relying solely on end-effectors (i.e., fingertips). However, state-of-the-art model-based planners for CRM are still focused on feasibility rather than optimality, limiting their ability to fully exploit CRM’s advantages.

We introduce a new paradigm that computes approximately optimal manipulator plans. This approach has two phases. Offline, we construct a graph of \emph{mutual reachable sets}, where each set contains all object orientations reachable from a starting object orientation and grasp. Online, we plan over this graph, effectively computing and sequencing local plans for globally optimized motion.

On a challenging, representative contact-rich task, our approach outperforms a leading  planner, reducing task cost by $61\%$. It also achieves a $91\%$ success rate across 250 queries and maintains sub-minute query times, ultimately demonstrating that globally optimized contact-rich manipulation is now practical for real-world tasks.
\end{abstract}





%
\IEEEpeerreviewmaketitle

\section{Introduction}

\begin{figure}[t]
    \centering
    \includegraphics[width=0.95\linewidth]{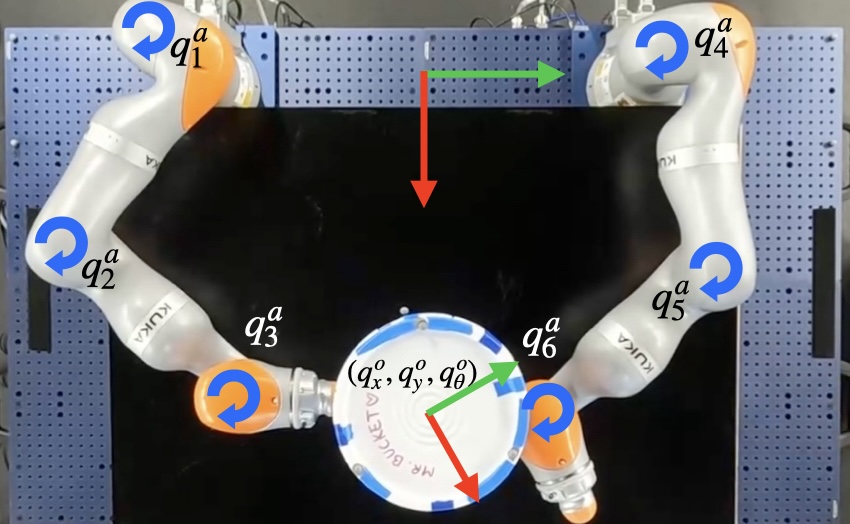}
    \caption{The bimanual KUKA iiwa-7 hardware setup with the cylindrical object. Because both the manipulators and the object are constrained to the $xy$-plane, only the three indicated joints are actuated on each arm. The object pose (shown) is defined with respect to the world frame located at the bimanual base (also shown).}
    \label{fig:system}
    \vspace{-5mm}
\end{figure}

\IEEEPARstart{R}{ecent} advances in large-scale behavior cloning (BC) have enabled robots to accomplish an unprecedented range of table-top tasks with remarkable success. However, due to limitations in teleoperation interfaces, BC is largely confined to end-effector manipulation, which struggles with large, bulky objects as well as small, delicate ones~\cite{black2024pi_0, lbmtri2025, chi2023diffusion, zhao2024aloha, team2024octo, shafiullah2023bringing, janner2022planning}. 
Handling such objects often requires contact-rich interactions between the manipuland and the manipulator or environment. For bulky objects (e.g., the cylinder in \Cref{fig:system}), bracing the object against the arms allows the robot to apply a wider range of wrenches, enabling more efficient plans. For small objects, multi-point support along the manipulator can provide the stability needed to achieve millimeter-level precision.

\textit{Contact-rich manipulation (CRM)}, which leverages the robot's entire surface for making contact with objects, remains a challenging problem for global planning. In contrast, there has been a spate of recent breakthroughs in \emph{local} planning for CRM (improving a trajectory given a feasible initialization), making it higher quality and more reliable. Most of this progress addresses the challenge posed by non-smooth, hybrid contact dynamics to gradient-based trajectory optimization methods. Some of these breakthroughs smooth the dynamics to enable gradient-based optimization~\cite{posa2014direct, sleiman2019contact, onol2019contact, onol2020tuning, kurtz2023inverse, pang2023global, suh2025ctr}, while others turn to sampling-based optimization~\cite{howell2022predictive, li2024drop}.

Despite this advancement in local planning, the global planning problem (computing a feasible and optimal path from scratch) remains underexplored. Existing global planners largely prioritize feasibility, with little consideration for optimality~\cite{pang2023global, cheng2022contact}. One reason optimality is so challenging is that the global planning search space is both high-dimensional and hybrid. In particular, the discrete component (typically contact-state or mode sequencing) is an intractable combinatorial problem. Our key insight is that for global path optimization to be tractable, we need to redefine this \textit{discrete decision space} to be more compact and tractable for global search. \textit{So, what is the right decision space for this problem?}  

Contact modes are one of the most low-level and common choices of discrete decision space. They specify which pairs of contact geometries are in contact and their relative velocities. Contact modes have been used with global optimization based on Mixed-Integer Programming (MIP)~\cite{marcucci2017approximate, marcucci2019mixed, graesdal2024towards, aceituno2020global, aceituno2022hierarchical}, sampling-based search~\cite{cheng2021contact, cheng2022contact}, A-star search~\cite{lee2015hierarchical}, and Monte-Carlo Tree Search (MCTS)~\cite{cheng2023enhancing, zhu2023efficient}. However, since the number of contact modes grows quickly with the number of contact points and degrees of freedom~\cite{huang2020efficient}, they can only be applied to the simplest of systems, like spherical point manipulators with a single polygonal object. 

Another common discrete decision space is hand-crafted motion primitives, such as ``push left'' \cite{hauser2010multi, lozano2014constraint, toussaint2018differentiable, zhu2023efficient}. Although such high-level discretization can drastically reduce the scale of the search, it can also make the planner too rigid, incapable of generating fine-grained motions. 

\begin{figure}[t]
    \centering
    \includegraphics[width=\linewidth]{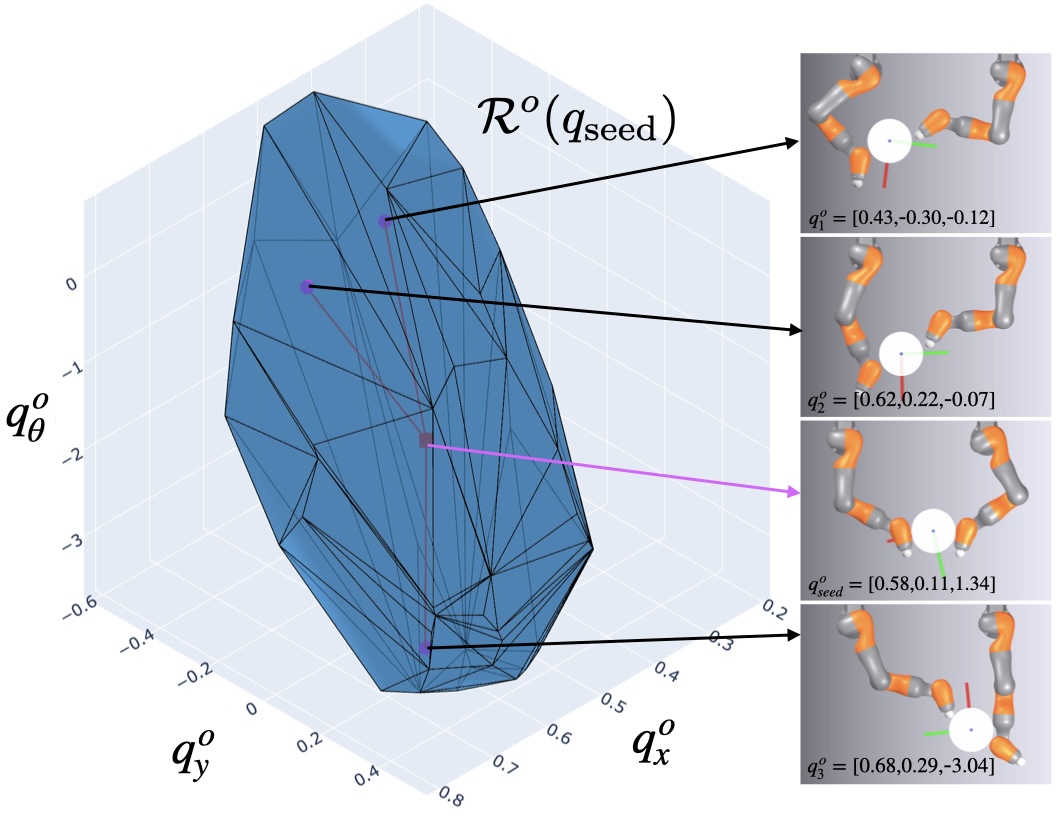}
    \caption{An illustration of different configurations ($q_{\text{seed}}$, $q_1$, $q_2$, $q_3$) inside an example MRS. Note how this single MRS encapsulates different contact modes (in $q_{\text{seed}}$, both wrists make contact; in $q_2$, a white end-effector makes contact; etc.). This is what makes MRS a useful discrete decision space - it abstracts away some of the combinatorial complexity of contact modes.
    }
    \label{fig:mrs_spans_multiple_modes}
    \vspace{-5mm}
\end{figure}

Our first contribution is a \textbf{novel, object-centric discrete decision space consisting of mutual reachable sets}. A mutual reachable set (MRS) comprises object states both forward and backward reachable in finite time from an initial object orientation and grasp. Roughly, our approach is to first cover object space with convex approximations of MRS, link them into a planning graph, and then apply recent optimal planning techniques for graphs of convex sets~\cite{marcucci2023shortest}. This produces an optimal object path, which is finally translated into a configuration path and a sequence of manipulator inputs.

We adopt an object-centric paradigm because object space is lower-dimensional than the full configuration (object and manipulator) space, making it feasible to cover with MRS. Object-centrism is also natural for contact-rich manipulation. In CRM, object motions are tightly coupled with manipulator motions, so object paths and initial grasps are highly descriptive on their own. 

The key advantage of the MRS decision space is that it strikes a balance between the expressiveness of low-level decision spaces (e.g., contact modes) and the efficiency of high-level ones (e.g., motion primitives). Each MRS aggregates many low-level contact modes into a single set (\Cref{fig:mrs_spans_multiple_modes}), substantially reducing the combinatorial complexity of global search. This reduction makes it possible to apply optimal search techniques and compute object-space plans in seconds (\Cref{sec:planning_experiments}).

At the same time, MRS are more expressive than fixed motion primitives. Each MRS represents a variety of different motions achievable by the local planner, and the decision-space granularity can be increased by adding more MRS as needed (\Cref{fig:n_mrs_tuning}). Together, these properties enable fast planning while maintaining broad coverage of feasible behaviors, resulting in high planning success rates across a wide range of queries (\Cref{sec:planning_experiments}).

Our second contribution is an \textbf{offline procedure to construct a graph of MRS}. In \Cref{sec:method_offline}, we describe algorithms for constructing convex-approximated MRS and producing an approximate cover of object space. Finally, we  define a planning graph over this set cover. We also introduce a graph cost-fitting feature that improves the correspondence between object-space and full-space planning.

Our third contribution is an \textbf{online hierarchical, multi-query planner} (\Cref{sec:method_online}). This planner operates in two stages: first, it uses the GCS algorithm~\cite{marcucci2023shortest} to produce an optimal object-space plan; second, it translates this plan into a full-space plan and manipulator input sequence by tracking the object plan with local planners. Three key features distinguish this planner: (1) \textit{approximately} optimal plans (empirically significantly better than SOTA), (2) sub-minute query times, and (3) high planning success rates across diverse queries.

While sampling-based motion planners for CRM also perform global search, none are asymptotically optimal in the CRM setting, and they often perform poorly in practice because their random growth strategies produce inefficient trajectories \cite{chavan2018hand, chen2021trajectotree, natarajan2023torque, pang2023global, venkatesh2025approximating}. Moreover, post-processing techniques that work well for sampling-based planners in collision-free settings are difficult to apply in the CRM setting. 

Furthermore, our planner is much more efficient than Reinforcement Learning (RL) methods, which search for global policies through uninformed search and extensive reward shaping \cite{kalashnikov2018scalable, andrychowicz2020learning, chen2022reorientation, zhou2023hacman, handa2023dextreme}. While RL methods can require years of simulation time, we can build our search graph in a few hours. 

In summary, our contributions are:
\begin{itemize}
    \item A novel, object-centric discrete decision space of mutual reachable sets (\Cref{sec:prelim:mrs})
    \item An offline framework for constructing a planning graph over a group of MRS (\Cref{sec:method_offline})
    \item A hierarchical, multi-query planner for use with this graph (\Cref{sec:method_online})
\end{itemize}
 
Plus, empirical results (\Cref{sec:planning_experiments}-\Cref{sec:hardware_experiments}) that demonstrate:
\begin{itemize}
    \item Applicability to challenging systems and tasks of real-world complexity 
    \item Significantly higher-quality plans than the SOTA sampling-based planner~\cite{suh2025ctr} (66\% cost improvement)
    \item Near-perfect planning success rates over a broad set of queries, highlighting robustness
    \item Sub-minute query times
\end{itemize}

\section{Preliminaries}\label{sec:prelim}

In this section, we begin by defining our problem setting formally. Then, we describe the local planners that we use as subroutines in our method. With all this preliminary information, we can then formally define mutual reachable sets, our object-centric decision space. Next, we review the GCS algorithm that we leverage to quickly compute optimal object paths over our graph of MRS. Finally, we wrap up by describing our challenging representative task.

\subsection{Problem Formulation}\label{sec:prelim:prob_formulation} 

We consider a discrete-time, quasi-dynamic \cite{mason2001mechanics} dynamical system of the form
\begin{equation}
\label{eq:prelim:dynamics} 
q_+ = f(q, u),
\end{equation}
where $q\coloneqq(q^a, q^o) \in \mathcal{Q} \subseteq \mathbb{R}^{n_a} \times \mathrm{SE}(2)$ is the system configuration, and $u$ the action or control input. We omit velocities due the quasi-dynamic assumption. The configuration $q$ is divided into the manipulator's actuated configurations $q^a \in \mathcal{Q}^a \subseteq \mathbb{R}^{n_a}$ and the unactuated object configurations $q^o \coloneqq (q^o_x, q^o_y, q^o_\theta) \in \mathcal{Q}^o \subseteq \mathrm{SE}(2)$ where $\mathcal{Q}^{o}$ is all object configurations that can be grasped (see \Cref{fig:system_illustration} for an illustration on an example). The action $u \in \mathcal{U} \subseteq \mathbb{R}^{n_a}$ consists of position commands tracked by a stiffness controller with diagonal gain matrix $\matr{K}_a \in \mathbb{R}^{n_a \times n_a}$.

Since we construct a planning graph over $\mathrm{SE}(2)$, we need to carefully define the distance function and set operations (inclusion, intersection, etc.) over this space. We view $\mathrm{SE}(2) \cong \mathbb{R}^2 \times S^1$ and equip it with the product Riemannian metric $g = \mathrm{d}p_x^2 + \mathrm{d}p_y^2 + w\,\mathrm{d}\theta^2$. The associated geodesic distance between $q_1^o = (p_1,\theta_1)$ and $q_2^o = (p_2,\theta_2)$ is
\begin{align}\label{eqn:dse2}
    d_{\mathrm{SE}(2)}(q_1^o,q_2^o)
    \coloneqq
    \sqrt{\|p_1 - p_2\|^2 + w \,\Delta\theta^2},
\end{align}
where $\Delta\theta = \mathrm{wrap}_{(-\pi,\pi]}(\theta_2 - \theta_1)$. We set $w$ to $1$. We take operations like equality, set inclusion, and set intersection to mean operations under this metric.

Next, we formally define our reorientation task. Our task assumes as input an object start pose $q^o_{\text{start}} \in \mathcal{Q}^o$, an object goal region 
\begin{align}
    \mathcal{Q}_{\text{goal}}^{o} = \{q^o \in \mathcal{Q}^{o}\mid d_\mathrm{SE(2)}(q^o,q_{\text{goal}}^{o}) \leq r \}
\end{align}\label{eq:prelim:goal_def} 
\noindent defined by goal state $q^o_{\text{goal}} \in \mathcal{Q}^o$ and tolerance $r \in \mathbb{R}^{+}$, horizon $T \in \mathbb{Z}^{+}$ and a cost function of the form 
\begin{align}
   c\left(q_{0:T}\right) = \sum_{t = 0}^{T-1} d\left(q_{t}, q_{t+1}\right)\label{eq:prelim:cost}
\end{align}
. We use task cost 
\begin{align}
d(q_t, q_{t+1}) = \norm{q^a_t - q^a_{t+1}}_2 
\label{eqn:our_task_cost}
\end{align}
throughout, which encourages velocity smoothness and the minimal path length in actuated space. However, the framework readily accommodates other task costs, such as deviation from a reference object trajectory, $d(q_t, q_{t+1}) = \norm{q^o_t - q^o_{t, \text{ref}}}_2$, or travel time, assuming a maximum joint velocity $v_{\text{max}}$, $d(q_t, q_{t+1}) = \norm{q^a_t - q^a_{t+1}}_{\infty}/v_{\text{max}}$.

Altogether, our task is to find configuration and control sequences $q_{0:T}, u_{0:T-1}$ that reach the goal region while minimizing the cost:

\begin{subequations}
\begin{align}
    \min_{q_{0:T},u_{0:T-1}} & c(q_{0:T}) \label{eqn:prelim:task_defn}\\
    \text{s.t.}\;\; & q_{0}^o = q^o_\text{start}, \\ 
                    & q_{t+1} = f(q_{t}, u_t),\;\; \forall t = 0, \ldots, T-1, \\ 
                    & q^{o}_{T} \in \mathcal{Q}^{o}_{goal}.
\end{align}
\end{subequations}

\subsection{Local Planners}\label{sec:prelim:planners}
We describe the local planners that we use as subroutines in our method: a contact-aware trajectory optimizer and a collision-free motion planner. The first kind of planner is used in the offline graph construction phase to find which states are reachable. Both kinds are used online in the hierarchical planner to translate the object path to a full-configuration space path. Following classical robotics literature \cite{koga1994multi}, we define two types of motions relevant to translation: \emph{transfer} motions, where the manipulator moves the object, and \emph{transit} motions, where the object is kept stationary during a regrasp. The contact-aware trajectory optimizer is used to translate transfer motions and the collision-free motion planner translates transit motions. 

First, we describe our contact-aware trajectory optimizer. Given an initial configuration $q$, an object goal configuration $q^o_{\text{goal}}$, and a horizon $T_{\pi}$, a contact-aware trajectory optimizer $\pi: \mathcal{Q} \times \mathcal{Q}^o \rightarrow \mathcal{U}^{T_{\pi}}$ takes in $(q, q^o_{\text{goal}})$ and produces $u_{0:T_{\pi}-1}$, a control trajectory which reaches $q^o_{\text{goal}}$ under the dynamics of \eqref{eq:prelim:dynamics}. Some popular ones are \cite{posa2014direct, alp-admm, howell2022dojo, suh2025ctr}. Based on our quasi-dynamic assumption, we choose~\cite{suh2025ctr}, which poses a trajectory optimization program subject to a convex, differentiable quasi-dynamics (CQDC) constraint. This is solved in the fashion of receding-horizon model predictive control (MPC); hence, we call it CQDC-MPC. CQDC-MPC also offers additional benefits, like numerical robustness, gradient-based contact-discovery via smoothing, and the ability to predict dynamics over longer timesteps. See \Cref{sec:appendix:ctr} for more details.

Next, we describe our collision-free motion planner. Given an initial configuration $q = (q^a, q^o)$, a desired robot configuration $q^a_\text{goal}$, and a horizon $T_{\psi}$, a collision-free motion planner $\psi : \mathcal{Q} \times \mathcal{Q}^a \to \mathcal{U}^{T_{\psi}}$ takes in $(q, q^a_\text{goal})$ and produces $u_{0:T_{\psi}-1}$, a control trajectory which brings the robot from $q^a$ to $q^a_\text{goal}$ without colliding with the stationary object at $q^o$.

Collision-free motion planning is a richly studied field, with planners such as \cite{marcucci2023motion} able to synthesize globally optimal collision-free trajectories in high-dimensional spaces. For this work, we choose the simple RRT-Connect planner \cite{kuffner2000rrt} and post-process with trajectory optimization. See \Cref{sec:appendix:collision_free} for details.

\subsection{Mutual Reachable Sets}\label{sec:prelim:mrs} 

\begin{figure}[t]
    \centering
    \includegraphics[width=0.9\linewidth]{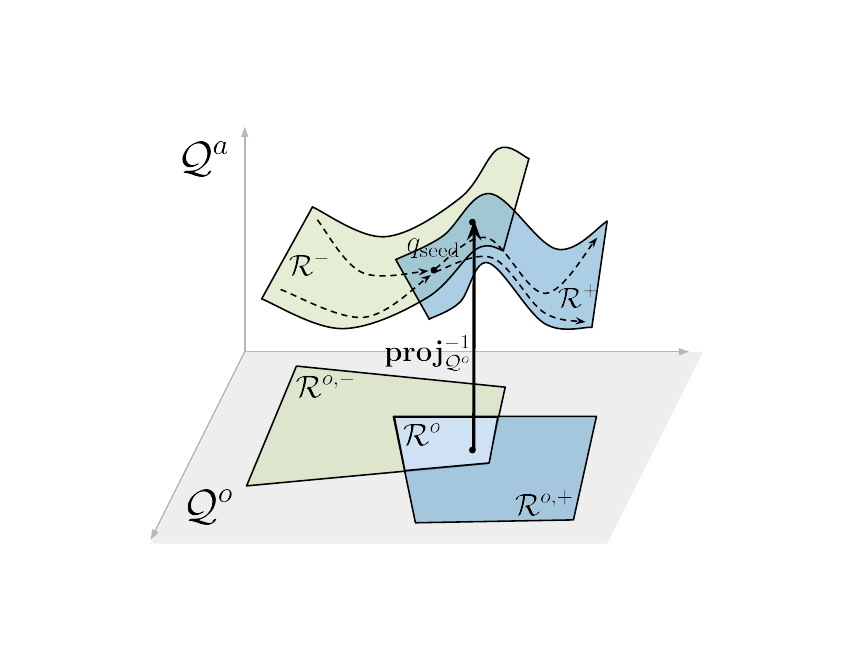}
    \caption{This figure shows the relationship between FRS $\mathcal{R}^{+} \in \mathcal{Q}$, BRS $\mathcal{R}^{-} \in \mathcal{Q}$, and MRS $\mathcal{R}^{o} \in \mathcal{Q}^{o}$. The FRS and BRS lie on contact manifolds in the full configuration space. They are defined by seed configuration $q_{seed}$, where they intersect, and the choice of contact-aware trajectory optimizer $\pi$. The MRS is defined in lower-dimensional object space, as the intersection of the projections of the FRS and BRS. We also illustrate the inverse projection operator, $\mathbf{proj}_{\mathcal{Q}^{o}}^{-1}$, that maps object states in the MRS to their full configurations.}
    \label{fig:frs_brs_mrs_diagram}
    \vspace{-5mm}
\end{figure}

Now we define the mutual reachable sets (MRS) of our contact-aware trajectory optimizer $\pi$, which serve as the basis of our framework. An MRS consists of all object configurations that are forward and backward reachable from a \textit{seed configuration} $q_{\text{seed}} \in \mathcal{Q}$ (a configuration where the manipulator and object are in contact) under $\pi$. The core innovation of our framework is that we take MRS to be our discrete elements, rather than contact modes or higher-level skills (e.g. rotating, pivoting). This choice of discretization reduces combinatorial complexity while still remaining fine-grained enough to solve general reorientation queries. 

We begin by defining \textit{forward} and \textit{backward reachable sets} (FRS, BRS). For convenience, we define $f_{T}: \mathcal{Q} \times  \mathcal{U}^{T} \rightarrow \mathcal{Q}$, which recursively applies dynamics for a length-$T$ control sequence $u_{0:T-1}$ from some initial state $q$. The finite-time FRS of a local planner $\pi$ is the set of all configurations that are reachable from $q_{\text{seed}}$ under $\pi$:
\begin{align}
    & \mathcal{R}^{+}_{\pi, T}(q_{\text{seed}}) = \{q \mid  q = f_{T}(q_{\text{seed}}, \pi(q_{\text{seed}}, q^o))\} \subseteq \mathcal{Q}.
\end{align}
Analogously, the finite-time BRS is all configurations that can reach $q_{\text{seed}}$ under $\pi$:
\begin{align}
    \mathcal{R}^{-}_{\pi, T}(q_{\text{seed}}) = \{q \mid  f_T(q, \pi(q, q^o_{\text{seed}})) = q_{\text{seed}}\} \subseteq \mathcal{Q}.
\end{align}
Though the FRS and BRS lie on contact manifolds (\Cref{fig:frs_brs_mrs_diagram}) in the full configuration space $\mathcal{Q}$, we are interested in their coordinate projections onto $\mathcal{Q}^o$, which we denote by a superscripted $o$: 
\begin{align}\label{eqn:proj_FRS_BRS}
\mathcal{R}^{o, +/-}_{\pi, T}(q_{\text{seed}}) \vcentcolon= \mathbf{proj}_{\mathcal{Q}^o}\left(\mathcal{R}^{+/-}_{\pi, T}(q_{\text{seed}})\right).
\end{align}

\noindent We are now ready to define the mutually reachable set. 
\begin{dfn}
\label{dfn:prelim:mrs}
\textbf{Mutually reachable set (MRS)} - given a seed configuration $q_{\text{seed}}$, planner $\pi$, and horizon $T$, the MRS is all object configurations that simultaneously can be reached from and can reach back to the seed in $2T$. 
\begin{align}
    \mathcal{R}_{\pi, T}^{o}(q_{\text{seed}}) \vcentcolon= \mathcal{R}^{o, +}_{\pi, T}(q_{\text{seed}}) \cap \mathcal{R}^{o, -}_{\pi, T}(q_{\text{seed}}) \subseteq \mathcal{Q}^o
\end{align}
\end{dfn}
\noindent See also \Cref{fig:frs_brs_mrs_diagram} for a visualization. For brevity, we refer to these sets henceforth as $\mathcal{R}^{+/-}, \mathcal{R}^{o, +/-}, \mathcal{R}^{o}$, dropping the dependency on $q_{\text{seed}}, \pi, T$.

\begin{lemma}\label{lemma:prelim:mutual_reachability}
Any two object states $q^o_1, q^o_2$ in the MRS are mutually reachable: $q^o_1$ is reachable from $q^o_2$ and vice versa.
\end{lemma}

See \Cref{sec:appendix:lemma1} for formal statement and proof. We will use this lemma later to show that our object plans are kinematically and dynamically feasible by construction, under some assumptions (\Cref{sec:method_online}). 

We also define an inverse projection operator that will be useful later (\Cref{sec:method_online}). Although each MRS $\mathcal{R}^{o}$ is defined in object space, it implicitly specifies a mapping between object configurations $q^{o} \in \mathcal{R}^{o}$ and corresponding full configurations $q \in \mathcal{Q}$. 
We define this as:
\begin{align}
\mathbf{proj}_{\mathcal{Q}^{o}}^{-1}(q^{o}, \mathcal{R}^{o}) = f_{T}(\mathcal{R}^{o}(q_{\text{seed}}), \pi(\mathcal{R}^{o}(q_{\text{seed}}), q^{o})).    
\end{align}\label{eqn:prelim:inverse_proj_operator}
where $\mathcal{R}^{o}(q_{\text{seed}})$ denotes the seed configuration of set $\mathcal{R}^{o}$. This means that the inverse projection is computed by reaching forward to $q^o$ from $\mathcal{R}^{o}(q_{\text{seed}})$ using $\pi$. 
See \Cref{fig:frs_brs_mrs_diagram} for an illustration.

\subsection{Shortest Paths on Graph of Convex Sets}\label{sec:prelim:gcs}
Offline, we use our local planner to cover object space with MRS and then connect them into a graph. At query-time, the first step of 
our hierarchical planner is to search for the shortest path through the MRS graph, yielding an object path, $q^{o}_{0:T}$. To find the shortest path on this graph of convex sets, we apply the Graph of Convex Sets (GCS) framework~\cite{marcucci2023shortest}. 

GCS takes in a directed graph $G = (\mathcal{V}, \mathcal{E})$ with vertex set $\mathcal{V}$ and edge set $\mathcal{E} \subseteq \mathcal{V}^2$. 
Each vertex $v \in \mathcal{V}$ is associated with a convex set  $\mathcal{X}_v$ and a nonnegative convex cost $\ell_v: \mathcal{X}_v \rightarrow \mathbb{R}^+$.
An element of $\mathcal{X}_v$ is denoted by $x_v \in \mathcal{X}_v$.
Similarly, each edge $e = (u, v) \in \mathcal{E}$ is associated with a Cartesian product of convex sets $\mathcal{X}_e \subseteq \mathcal{X}_u \times \mathcal{X}_v$ and a nonnegative convex cost $\ell_e: \mathcal{X}_u \times \mathcal{X}_v \rightarrow \mathbb{R}^+$.
An element of $\mathcal{X}_e$ is denoted by $(x_u, x_v) \in \mathcal{X}_e$.
Next, we define a path $p$ as a sequence of distinct vertices that connect a source $s \in \mathcal{V}$ to a target $t \in \mathcal{V}$, with $\mathcal{E}_p$ as the edges traversed by $p$. Let $\mathcal{P}$ be the set of all such paths. 
The shortest path is the solution to the following optimization problem:
\begin{subequations}
    \begin{align}
        \min_{p \in \mathcal{P}}  \quad &\sum_{v \in p}\ell_{v}(x_v) + \sum_{e = (u,v) \in \mathcal{E}_p} \ell_e(x_u, x_v), \\
        \text{s.t.} \;\; & x_v \in\mathcal{X}_v ,\;\; \forall v \in p, \\
        & (x_u, x_v) \in \mathcal{X}_e, \;\; \forall e = (u,v) \in \mathcal{E}_p.
    \end{align}\label{eq:prelim:GCS}
\end{subequations}
This shortest paths problem can be transcribed into a mixed-integer convex program (MICP). While the MICP program can be solved to optimality using MIP solvers, the GCS framework introduces a tight \textit{convex relaxation} that can be solved much faster, while yielding effectively optimal results in many cases. We adopt the GCS framework, which enables us to find optimal object paths in seconds. Also, we verify later that the optimality gap with MIP is negligible (\Cref{sec:planning_experiments:quality}).  

\subsection{Task Specification}\label{sec:task_specification} 
Although the planner we propose is not limited to a particular robotic system, we introduce our experimental setup here to ground subsequent discussions in a concrete context.

\begin{figure}[h!]
\centering
\subfloat{
    \includegraphics[width=0.45\columnwidth]{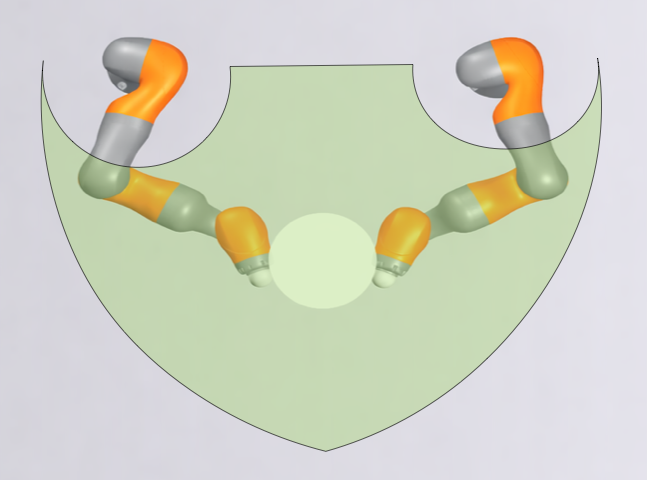}
}
\hfill
\subfloat{
    \includegraphics[width=0.45\columnwidth]{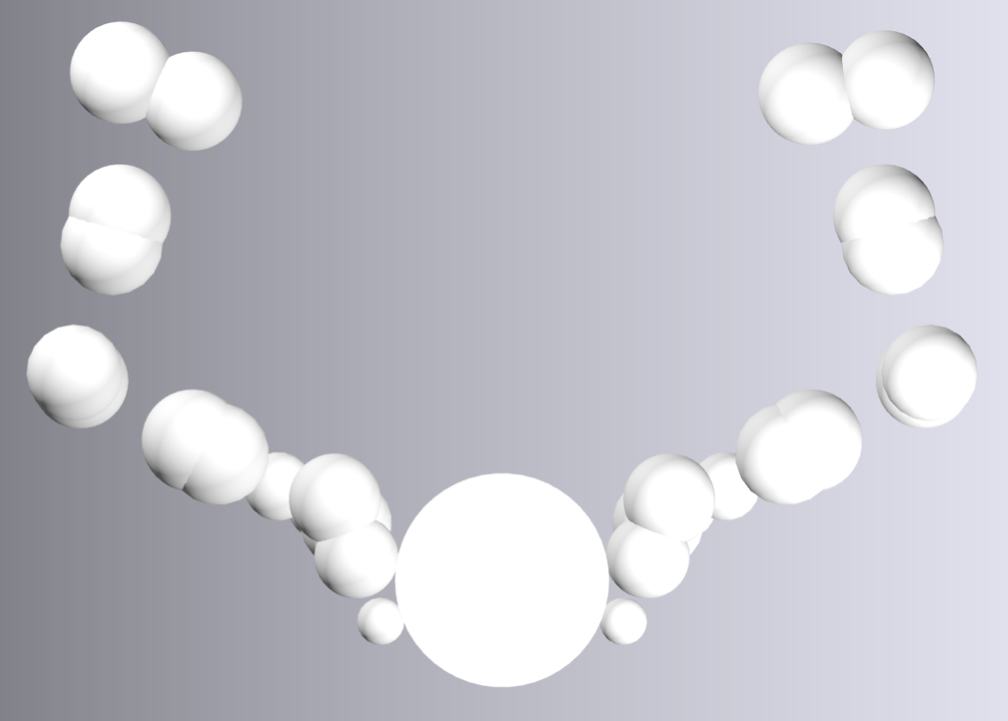}
}
\caption{Left: visual geometry of the bimanual KUKA iiwa-7 system, with the object workspace $\mathcal{Q}^{o}$ overlaid in green. $\mathcal{Q}^{o}$ describes all object positions that can be grasped. Right: collision model used for planning. Each arm has 14 spheres.
} 
\label{fig:system_illustration}
\vskip -0.1 true in
\end{figure}

We consider a bimanual KUKA iiwa system (\Cref{fig:system}), where each arm has 7 joints but only 3 (shoulder, elbow, wrist) are actuated to constrain motion in the $xy$-plane. The task is to move a cylindrical object to a desired pose. The system has 3 unactuated DOFs, 6 actuated DOFs ($n_a=6$), and 29 collision geometries. We assume known geometric models of both the robot and the object, which are shown in \Cref{fig:system_illustration}.

This is a challenging and representative contact-rich manipulation task: it requires exploiting intrinsic dexterity to complete efficiently, while reasoning about multi-point contacts and complex arm geometries. Compared to the simpler end-effector or gripper systems typically studied in model-based planning for contact-rich manipultion, this setup is significantly higher dimensional and better tests the capabilities of our planner.

\section{Methodology - Offline Graph Construction}\label{sec:method_offline}

We now present the offline phase of our framework, in which we construct a planning graph over MRS. We first describe how to compute a single MRS and its convex approximation.
We then specify a simple algorithm for covering the object space with these convex-approximated MRS. Finally, we explain how we define a graph over this group of sets. In the following \Cref{sec:method_online}, we will describe the online phase, which uses this graph at query-time to produce a control sequence. 

\subsection{Computing a Convex-Approximated MRS}\label{sec:method:offline:reachset}

\begin{figure}[t]
    \centering
    \includegraphics[width=\linewidth]{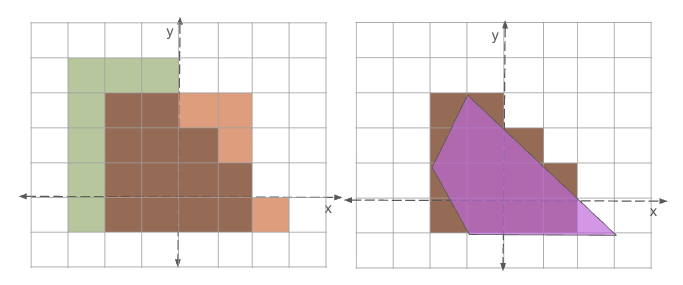}
    \caption{
    Demonstration of computing a convex-approximated MRS on a toy example in 2D object space. Left: after we compute the projected FRS $\mathcal{R}^{o, +}_{\Delta}$ (orange) and BRS $\mathcal{R}^{o, -}_{\Delta}$ (green), we intersect them to get the discrete MRS $\mathcal{R}^{o}_{\Delta}$ (brown). Right: next, we find a convex approximation of $\mathcal{R}^{o}_{\Delta}$. The figure shows a likely convex approximation produced by the algorithm IRIS-ZO: $\hat{\mathcal{R}}^{o}_{\Delta}$ (violet). It lies mostly within $\mathcal{R}^{o}_{\Delta}$, but may exceed it somewhat. We want to avoid such ``overapproximation'' as much as possible, since it amounts to introducing unreachable sets into our reachable set approximation. }
    \label{fig:iris_example_2d}
\end{figure}

\begin{algorithm}[t]
\caption{\small{\texttt{ComputeConvexApproximatedMRS}}}\label{alg:mrs}
\KwIn{Seed grasp $q_\text{seed}\in\mathcal{Q}$, resolution $\Delta\in\mathbb{R}^{+}$}
\KwOut{Convex-approximated MRS $\hat{\mathcal{R}}_{\Delta}^{o}$}
$\mathcal{R}^{o,+}_{\Delta} \leftarrow \texttt{set()}$\;
$\mathcal{R}^{o,-}_{\Delta} \leftarrow \texttt{set()}$\;
$\texttt{grid} \leftarrow \texttt{Discretize}(\mathcal{Q}^o, \Delta)$\;
\For{$\texttt{cell} \in \texttt{grid}$}{
  $q^o_\text{cell} \leftarrow \texttt{GetCenter}(\texttt{cell})$\;
  $q_\text{forward} \leftarrow f_{T}(q_{\text{seed}}, \pi(q_{\text{seed}}, q^o_\text{cell}))$\;
  \If{$q^o_\text{forward} \in \texttt{cell}$}{
    $\mathcal{R}^{o,+}_\Delta.\texttt{add}(\texttt{cell})$\;
    $q_\text{backward} \leftarrow f_{T}(q_{\text{forward}}, \pi(q_{\text{forward}}, q^o_\text{seed}))$\;
    \If{$d_{\mathrm{SE}(2)}(q^o_{\text{backward}}, q^o_\text{seed}) \le \text{threshold}$}{
      $\mathcal{R}^{o,-}_\Delta.\texttt{add}(\texttt{cell})$\;
    }
  }
}
$\mathcal{R}^{o}_{\Delta} \leftarrow \mathcal{R}^{o,+}_{\Delta} \cap \mathcal{R}^{o,-}_{\Delta}$\;
$\hat{\mathcal{R}}^{o}_{\Delta} \leftarrow \texttt{IrisZo}(\mathcal{R}^{o}_{\Delta})$\;
\Return $\hat{\mathcal{R}}^{o}_{\Delta}$\;
\end{algorithm}

We first describe the process to compute a single MRS for a given seed grasp $q_{\text{seed}}$ (\Cref{alg:mrs}, \Cref{fig:iris_example_2d}). We begin by discretizing the object space into a grid with resolution $\Delta$. We then compute the projected FRS and BRS (\Cref{eqn:proj_FRS_BRS}) on this discretized object space as binary occupancy maps, denoting them $\mathcal{R}^{o, +}_{\Delta}$ and  $\mathcal{R}^{o, -}_{\Delta}$. Specifically, to compute the $\mathcal{R}^{o, +}_{\Delta}$, we attempt to reach each grid cell from $q_{\text{seed}}$. Similarly, for $\mathcal{R}^{o, -}_{\Delta}$, we attempt to reach $q_{\text{seed}}$ from each grid cell. Then, the MRS is obtained by their intersection: $\mathcal{R}^{o}_{\Delta} = \mathcal{R}^{o, +}_{\Delta} \cap \mathcal{R}^{o, -}_{\Delta}$.

Since an MRS is generally non-convex, we need to approximate it with a convex set, $\hat{\mathcal{R}}^{o}_{\Delta}$, to use it within the GCS framework\footnote{Since our convex-approximated sets $\hat{\mathcal{R}}^o_\Delta$ belong to $\mathrm{SE}(2)$, which is non-Euclidean, the notion of convexity requires precise mathematical justification. See \Cref{sec:appendix:se2_convexity}.}. We can find a convex polytope that is approximately contained in the MRS using IRIS-ZO \cite{werner2024faster}, a fast stochastic algorithm. IRIS-ZO guarantees the resulting $\hat{\mathcal{R}}^{o}_{\Delta}$ to be $\epsilon$-correct, where correctness refers to the volume of $\hat{\mathcal{R}}^{o}_{\Delta}$ exceeding $\mathcal{R}^{o}_{\Delta}$ divided by the volume of $\mathcal{R}^{o}_{\Delta}$, and $\epsilon$ can be specified. The $\epsilon$-correctness is desirable, as introducing unreachable states into $\hat{\mathcal{R}}^{o}_{\Delta}$ adversely affects planning (\Cref{sec:method_online}).

\subsection{Covering Object Space with MRS}\label{sec:method:offline:covering}

\begin{figure}[h!]
\centering
\subfloat[5 MRS]{
    \includegraphics[width=0.45\columnwidth]{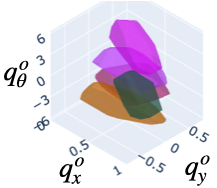}
}
\hfill
\subfloat[10 MRS]{
    \includegraphics[width=0.45\columnwidth]{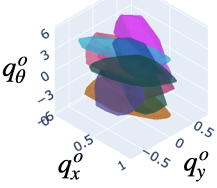}
}
\vspace{2pt}
\subfloat[15 MRS]{
    \includegraphics[width=0.45\columnwidth]{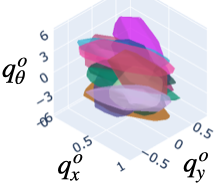}
}
\hfill
\subfloat[25 MRS]{
    \includegraphics[width=0.45\columnwidth]{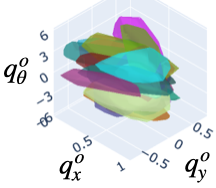}
    \label{fig:set_cover:25_sets}
}
\caption{Illustration of object space covering algorithm (\Cref{alg:cover}) in $\mathrm{SE}(2)$ at different iterations.
} 
\label{fig:set_cover}
\vskip -0.1 true in
\end{figure}

\begin{algorithm}[t]
\caption{\small{\texttt{ComputeObjectSpaceCover}}}\label{alg:cover}
\KwIn{Cover fraction $\alpha \in [0,1)$, resolution $\Delta \in \mathbb{R}$}
\KwOut{Collection of sets $C = \{\hat{\mathcal{R}}_{\Delta}^{o,i}\}_{i=1}^{\lvert C \rvert}$}
$C \gets \texttt{collection}()$\;
\While{$\dfrac{\mathrm{vol}\left(\bigcup_{i=1}^{\lvert C \rvert} \hat{\mathcal{R}}^{o,i}_\Delta\right)}{\mathrm{vol}(\mathcal{Q}^o)} \le \alpha$}{
    \tcp{sample uncovered} 
    $q^{o}_\text{seed} \sim \mathrm{Uniform} \left(\mathcal{Q}^{o} \setminus \bigcup_{i=1}^{\lvert C \rvert} \hat{\mathcal{R}}^{o,i}_\Delta\right)$\; 
    $q_\text{seed} \gets \texttt{GenerateGrasp}(q^o_\text{seed})$\;
    $\hat{\mathcal{R}}^{o, \lvert C \rvert + 1}_\Delta \gets \texttt{ComputeConvexApproximatedMRS}(q_\text{seed}, \Delta)$\;
    $C.\texttt{add}\left(\hat{\mathcal{R}}^{o, \lvert C \rvert + 1}_\Delta\right)$\;
}
\KwRet $C$\;
\end{algorithm}

Given the ability to compute an approximate MRS from a single seed, we now seek to build an $\alpha$-approximate cover of the object space $\mathcal{Q}^{o}$.

\begin{dfn}\label{definition:alpha_approx_cover}
Following~\cite{werner2024faster}, an \textbf{$\alpha$-approximate cover} of $Q^{o}$ is a collection of convex sets $\hat{\mathcal{R}}_{\Delta}^{o, 1},\ldots,\hat{\mathcal{R}}_{\Delta}^{o, N} \subseteq Q^{o}$ whose union covers at least an $\alpha$-fraction of its volume:
\begin{equation}
    \mathrm{vol}\left( 
    \overset{N}{ \underset{i=1}{\cup}}\hat{\mathcal{R}}_{\Delta}^{o, i}
    \right) \geq \alpha \cdot \mathrm{vol}(Q^{o})
\end{equation}
\end{dfn}

The closer $\alpha$ is to 1, the better: this increases the likelihood that a query $(q^{o}_{\text{start}}, q^{o}_{\text{goal}})$ is contained in some MRS and that our planner can be run. Higher $\alpha$ leads to higher planning success rates.  

We construct the cover iteratively (\Cref{alg:cover}, \Cref{fig:set_cover}). At each iteration, we sample the uncovered region of $\mathcal{Q}^{o}$ to obtain a new seed object configuration $q^o_{\text{seed}}$. We then generate a grasp for $q^o_{\text{seed}}$, yielding a full configuration $q_{\text{seed}}$. The grasp generation technique is system and task specific and detailed in \Cref{sec:appendix:qseed}. From this $q_{\text{seed}}$, we construct an approximate MRS (\Cref{alg:mrs}). Under mild assumptions, this procedure is probabilistically complete (see \Cref{sec:appendix:cover} for proof). Adding more sets can monotonically increase coverage of $\mathcal{Q}^{o}$, converging to full coverage. 

\subsection{Linking MRS Into a Graph}\label{sec:method:offline:linking}

\begin{figure}[h!]
\centering
\subfloat[Toy example]{
    \includegraphics[width=0.45\columnwidth]{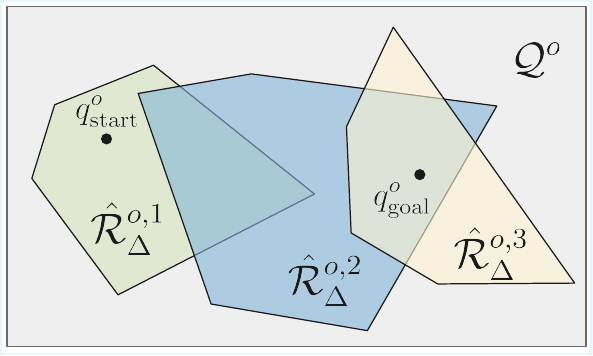}
    \label{fig:construct_gcs_graph_A}
}
\hfill
\subfloat[GCS graph]{
    \includegraphics[width=0.45\columnwidth]{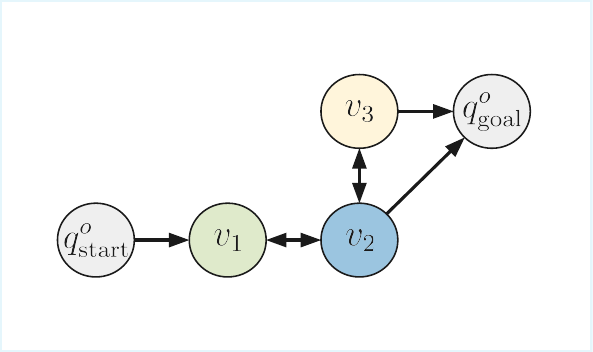}
    \label{fig:construct_gcs_graph_B}
}
\vspace{2pt}
\subfloat[Online: Stage 1]{
    \includegraphics[width=0.45\columnwidth]{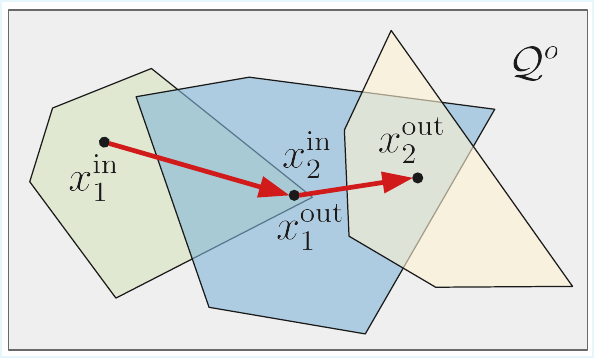}
    \label{fig:construct_gcs_graph_C}
}
\hfill
\subfloat[Online: Stage 1 in GCS graph]{
    \includegraphics[width=0.45\columnwidth]{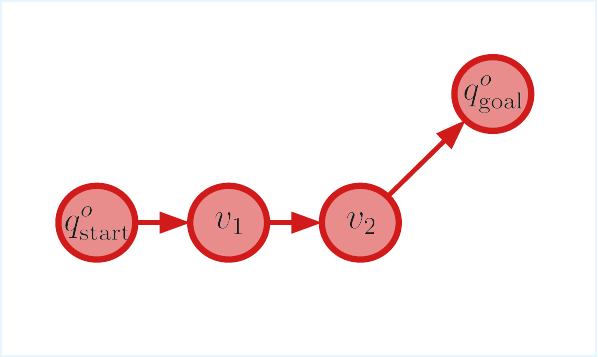}
    \label{fig:construct_gcs_graph_D}
}
\vspace{2pt}
\subfloat[Online: Stage 2, first steps]{
    \includegraphics[width=0.45\columnwidth]{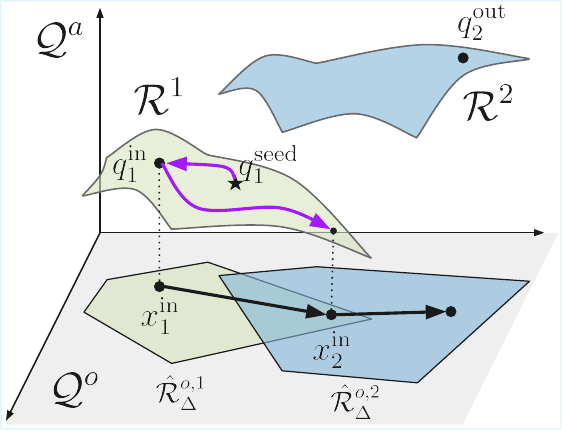}
    \label{fig:construct_gcs_graph_E}
}
\hfill
\subfloat[Online: Stage 2, next steps]{
    \includegraphics[width=0.45\columnwidth]{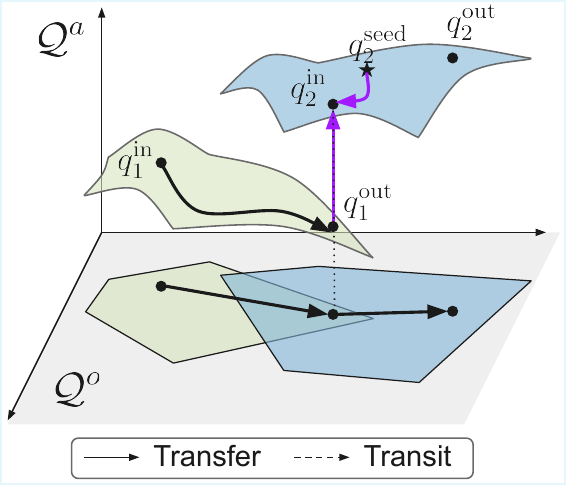}
    \label{fig:construct_gcs_graph_F}
}
\caption{Illustration of online planning with a toy example. First row: example of 3 MRS in 2D object space. 
Second row: show the first stage of hierarchical planning, producing an optimal object plan. It consists of an MRS sequence (\Cref{fig:construct_gcs_graph_D}) and a continuous object path  (\Cref{fig:construct_gcs_graph_C}).
Third row: shows the second stage of hierarchicla planner, translating the object plan to manipulator inputs. In~\Cref{fig:construct_gcs_graph_E}, shown in purple, we compute the initial grasp with $\mathbf{proj}^{-1}_{\mathcal{Q}^{o}}$ and then track the first transfer segment $(x^{\text{in}}_1, x^{\text{in}}_{2})$ with $\pi$. In~\Cref{fig:construct_gcs_graph_F}, we find the next grasp with $\mathbf{proj}^{-1}_{\mathcal{Q}^{o}}$ and then compute the transit with $\psi$.
} 
\label{fig:construct_gcs_graph}
\vskip -0.1 true in
\end{figure}

We shorten the set notation from $\hat{\mathcal{R}}^{o, i}_{\Delta}$ to $\mathcal{R}^{i}$. Supposing we have $N$ such sets $\{\mathcal{R}^{i}\}_{i=1}^{N}$ from the prior section, we can now define a planning graph over them. See the first row in~\Cref{fig:construct_gcs_graph} for a illustrated example. 
First, recall that we consider two types of motion (\Cref{sec:prelim:planners}): ``transfer'' motions, where the manipulator moves the object, and ``transit'' motions, where the object is kept stationary during a regrasp. We handle these motions differently when building the graph.   

In order to build a smaller GCS graph with fewer edges, we represent transfer motions as graph vertices, rather than edges. Namely, for each convex set $\mathcal{R}^{i}$, we define abstract vertex $v_i$ with continuous state $x_{v_i} = \begin{bmatrix} x_{v_{i}}^{\text{in}} & x_{v_{i}}^{\text{out}} \end{bmatrix}^T \in \mathcal{R}^{i} \times \mathcal{R}^{i}$. $ x_{v_{i}}^{\text{in}}$ and $x_{v_{i}}^{\text{out}}$ are interpreted as the endpoints of a motion segment inside $\mathcal{R}^{i}$. The associated vertex cost $\ell_{v_i}(x_{v_i})$ represents the task cost (\Cref{eq:prelim:cost}) of moving the object from $x_{v_i}^{\text{in}}$ to $x_{v_i}^{\text{out}}$ using $\pi$.

For transit motions, we assume a regrasp is possible between any two configurations $q_i, q_j$ with the same object configuration ($q_i^{o} = q_j^{o}$). This means that for all intersecting MRS $\mathcal{R}^i$ and $\mathcal{R}^j$, we add two directed edges. We add an edge $e_{ij} = (v_i, v_j)$, with constraint $x^\text{out}_{v_i} = x^\text{in}_{v_j}$. The associated edge cost $\ell_{e_{ij}}(x_{v_i}, x_{v_j})$ represents the task cost (\Cref{eq:prelim:cost}) of moving the system from $\mathbf{proj}^{-1}_{\mathcal{Q}^o}(x^{\text{out}}_{v_i}, \mathcal{R}^i)$ to $\mathbf{proj}^{-1}_{\mathcal{Q}^o} (x^{\text{in}}_{v_j}, \mathcal{R}^j)$ using $\psi$. The second edge $e_{ji}$ is defined analogously.

Finally, given a query $(q^{o}_{\text{start}}, q^{o}_{\text{goal}})$, we can incorporate it into the graph by adding singleton sets $v_\text{start}\coloneqq\{q^{o}_{\text{start}}\}$ and $v_\text{goal}\coloneqq\{q^{o}_{\text{goal}}\}$ to the graph and connecting them to any set which contains them.
Namely, we create zero-cost edges $e_{\text{start}, i} = (v_{\text{start}}, v_i)$ for all $i$ such that $q^{o}_{\text{start}} \in \mathcal{R}^{i}$, and $e_{i, \text{goal}} = (v_i, v_{\text{goal}})$ for all $i$ such that $q^{o}_{\text{goal}} \in \mathcal{R}^{i}$. 

\subsection{Fitting Surrogate GCS Vertex and Edge Costs}\label{sec:method:offline:cost_fitting} 

A key design question is how to assign costs to GCS vertices and edges.

Each GCS vertex and edge specifies only the start and goal configurations of a transit or transfer motion. To evaluate their corresponding task costs, we must therefore run a local planner ($\pi, \psi$) to generate a physically realizable trajectory between these configurations. As a result, vertex and edge task costs are only available as \emph{black-box} functions, since evaluating them requires executing a planner.

However, GCS requires vertex and edge costs to be non-negative and convex. To satisfy these requirements, we fit non-negative, convex surrogate cost functions to samples of the black-box task cost. This fitting is performed independently for each edge using least-squares regression (see \Cref{sec:appendix:cost} for details).
   
\begin{figure}[h!]
\centering
\subfloat[Stage 1: object-space planning over MRS graph]{
    \includegraphics[width=0.45\textwidth]{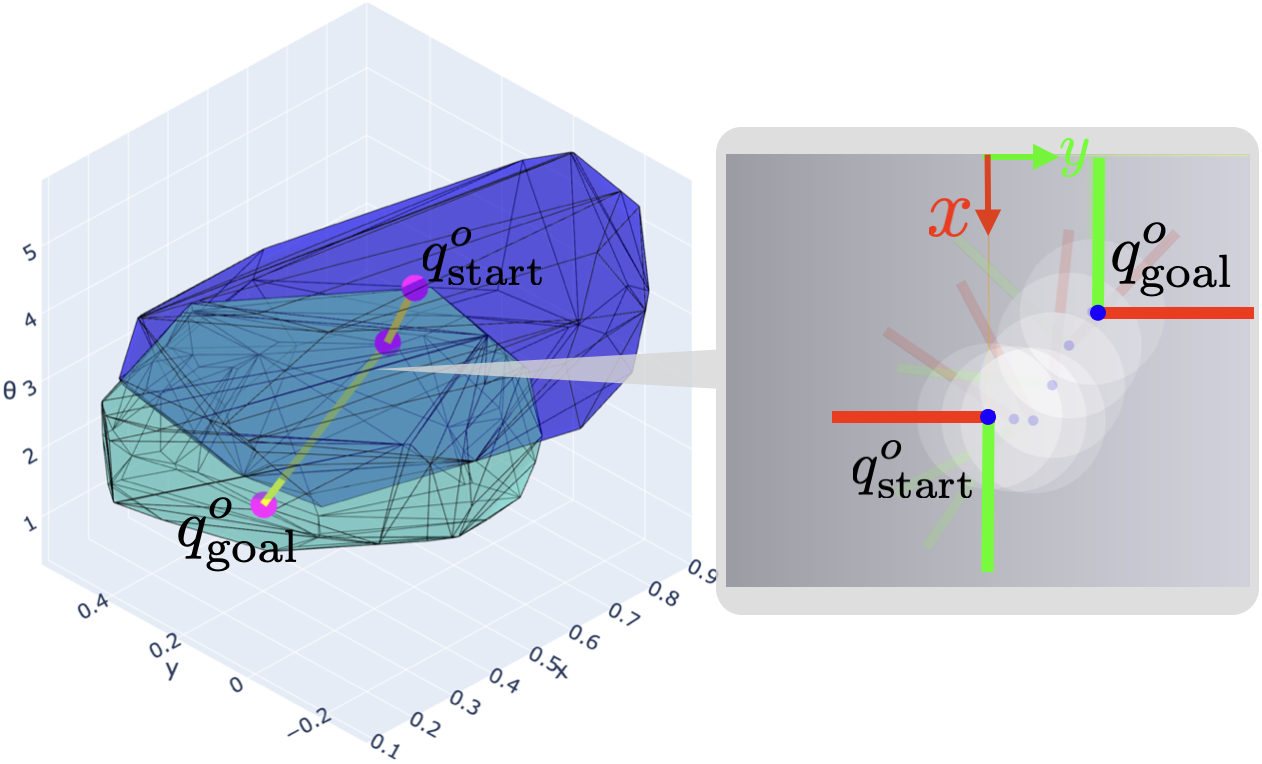}
    \label{fig:online_wide_figure:stage1}
}
\hfill
\subfloat[Stage 2: translation to full configurations and inputs]{
    \includegraphics[width=0.42\textwidth]{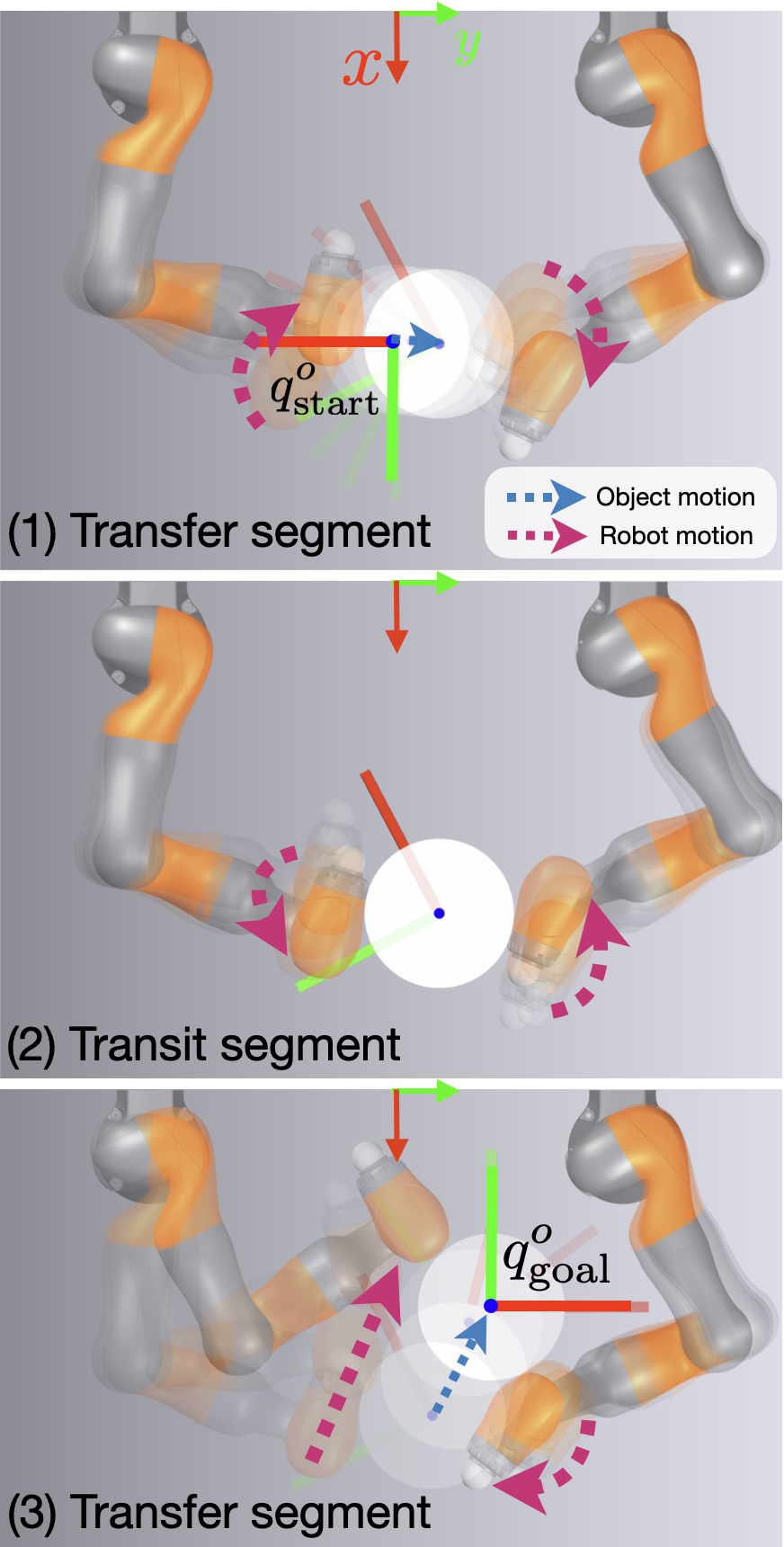}
    \label{fig:online_wide_figure:stage2}
}
\caption{
Illustration of online hierarchical planning.
}
\label{fig:online_wide_figure}
\vskip -0.1 true in
\end{figure}

\section{Methodology - Online Hierarchical Planning}\label{sec:method_online}

In the previous section, we described an offline procedure for constructing a planning graph of mutually reachable sets in object space. We now introduce the hierarchical, multi-query planner that uses this graph to  compute plans. The planner operates in two stages: (1) computing an object-space plan over the graph using GCS, and (2) translating this object plan into a full configuration (object and manipulator) plan and manipulator control sequence. This design has three advantages: approximately optimal plans, sub-minute query times, and high planning success rates across diverse queries.

Firstly, the planner produces optimal object paths and \textit{approximately optimal} full-space plans. While it is difficult to formally bound the suboptimality of the full-space plans, experiments show a 66\% reduction in task cost compared to the SOTA baseline. This suggests that object-space optimization is an effective proxy for full-space planning. 

Secondly, the planner achieves fast query times by performing the computationally expensive discrete search in low-dimensional object space rather than full configuration space. This is sensible, as much of the full space is redundant or irrelevant to manipulation tasks. The GCS search typically completes within a second (\Cref{sec:planning_expexperiments:runtimes}), and the remaining translation steps rely on contact-aware trajectory optimization and collision-free motion planning, which are both well-established and efficient routines. Hence, overall, the planner can obtain sub-minute query times. 

Thirdly, the planner achieves a high planning success rate, reliably producing plans that are kinematically and dynamically feasible under the dynamics \eqref{eq:prelim:dynamics}. A common limitation of hierarchical planners is that high-level plans often become infeasible at lower levels, as the high-level typically ignores kinematic and dynamic constraints. Our approach avoids this issue by design. To see this, recall that object plans contain both transfer and transit segments. Now, transfer segments are within reachable sets, which, by definition, correspond to physically feasible motions 
(\Cref{lemma:prelim:mutual_reachability}). Next, we always assume transit segments are feasible (\Cref{sec:method:offline:linking}). This is often a safe assumption in $\mathrm{SE}(2)$, since a regrasp can be performed by letting go of the object and finding a collision-free path to the next grasp.\footnote{In $\mathrm{SE}(3)$, there may be an additional challenge of keeping the object stationary during a regrasp.} Hence, our object plans are largely translatable to valid manipulator plans.

Overall, this design provides a fast and dependable framework for global planning in contact-rich manipulation. We describe the two stages of planning in detail below.

\subsection{Stage 1: Object-Space Planning Over the MRS Graph}
\label{sec:method:online:stage1}

Given a query $(q_\text{start}^o, q_\text{goal}^o)$, we first augment the precomputed MRS graph with the start and goal nodes, then run GCS to solve for an optimal object trajectory $q^{o}_{0:\tau}$ along with the associated sequence of reachable sets $\mathcal{R}^{0:\tau-1}$. See \Cref{fig:construct_gcs_graph_C} and \Cref{fig:construct_gcs_graph_D} for an illustration on a toy example.

\subsection{Stage 2: Translation to Full Configurations and Inputs}
\label{sec:method:online:stage2}

\begin{algorithm}[t]
\caption{\small{Transcription of GCS plan}}
\textbf{Input:} Object trajectory $q_{0:\tau}^{o}$, sequence of sets $\mathcal{R}^{0:\tau-1}$\; 
\textbf{Output:} Control trajectory ${u}_{0:\tau^{'}}$\;
\tcc{Initialize state and output}
$q \leftarrow \mathbf{proj}^{-1}_{\mathcal{R}^0}(q_{0}^{o})$\;
$u_{\text{traj}} \leftarrow \texttt{list()}$\;
%
%
\For {$t = 0,\dots,\tau-2$} { \label{alg:trajopt_ctr:while}
    \tcc{Generate contact segment}
    $u_{0:\tau_{\pi}} \leftarrow \pi(q,q^{o}_{t+1})$\;
    $u_{\text{traj}}.$\texttt{extend}$(u_{0:\tau_{\pi}})$\;
    \tcc{Generate grasp for next set}
    $q' \leftarrow \mathbf{proj}^{-1}_{\mathcal{R}^{t+1}}(q^{o}_{t+2})$\;
    \tcc{Generate regrasp segment}
    $u_{0:\tau_{\psi}} \leftarrow \psi(q,q')$\;
    $u_{\text{traj}}.$\texttt{extend}$(u_{0:\tau_{\psi}})$\;
    \tcc{Update current state}
    $q \leftarrow q'$\;
}
\tcc{Generate final contact segment}
$u_{0:\tau_{\pi}} \leftarrow \pi(q,q^o_{\tau})$\;
$u_{\text{traj}}.$\texttt{extend}$(u_{0:\tau_{\pi}})$\;
\Return $u_{\text{traj}}$
\end{algorithm}\label{alg:methodology:online:translate}

Given the object-space plan from Stage~1, this stage computes the corresponding full configuration space path $q_{0:\tau^{'}}$ and control sequence $u_{0:\tau^{'}}$. As outlined in \Cref{alg:methodology:online:translate}, the procedure alternates between transfer (moving the object within an MRS) and transit (regrasping between MRS).
During each transfer phase, we use the contact-aware trajectory optimizer $\pi$ (CQDC-MPC) to track the object path by moving the object using the manipulator. For the transit phase, the new grasp is computed using the inverse projection operator defined in \Cref{eqn:prelim:inverse_proj_operator} 
and a collision-free path to this grasp is generated using $\psi$ (BiRRT). See \Cref{fig:construct_gcs_graph_E} and \Cref{fig:construct_gcs_graph_F} for an illustration on a toy example.

\subsection{Optimality}\label{sec:method:online:sources_of_failure}\label{sec:method:online:optimality}
Overall, the object path found by GCS is effectively optimal with respect to the fitted task cost, up to the choice of set cover and continuous path parametrization. As shown in \Cref{sec:planning_experiments:quality}, the suboptimality gap between GCS and the optimal MIP solution is negligible. Formal guarantees for the corresponding configuration-space path are harder to make, since GCS optimizes an approximation of the true task cost (\Cref{sec:method:offline:cost_fitting}). Nonetheless, our experiments show that the resulting configuration-space paths consistently outperform a SOTA baseline (\Cref{sec:planning_experiments:quality}).

\subsection{Sources of Planning Failure}\label{sec:method:online:sources_of_failure}
We now list the main sources of planning failure, 
which generally stem from imperfections in the offline graph construction.
\begin{enumerate}
\item In Stage 1, the start or goal object poses ($q^o_\text{start}$ or $q^o_\text{goal}$) may fall outside the graph’s coverage, even if the manipulator can kinematically reach them. This occurs because the graph only provides an \textit{approximate} cover of the object workspace.
\item In Stage 2, translation of a transfer segment may fail if the target object pose is kinematically or dynamically unreachable. This can happen because the convex-approximated MRS include some unreachable states due to discretization and convexification errors.
\item In Stage 2, grasp generation may fail for similar reasons: the object pose of the regrasp target configuration may be unreachable. 
\item In Stage 2, translation of a transit segment may fail when no collision-free regrasp path exists. While our planner assumes such a path is always available, occasional failures occur when the manipulators become trapped around the object or the environment geometry before the intended regrasp (\Cref{fig:transit_failure}).
\end{enumerate}

\subsection{Sampling Multiple Object Paths to Mitigate Failure}\label{sec:method:online:sampling}
Since all failure modes except the first can be avoided with a different choice of object path, we address them altogether by sampling multiple candidate object paths in Stage 1. While GCS normally returns the single lowest-cost object path, it also provides traversal probabilities for each edge in the optimal flow (the dual of the shortest-path problem). These probabilities let us sample additional suboptimal paths, with sampling likelihood inversely proportional to their task cost. We then translate all sampled object paths through Stage 2 and select the lowest-cost feasible one. This strategy is highly parallelizable, allowing us to notably improve success rates without significantly increasing wall-clock time (see ablation study in \Cref{sec:ablation:path_sampling}). 

\section{Planning Experiments}\label{sec:planning_experiments}
In this and the following two sections, we present our experimental results. We begin here by comparing our approach, Graph of Reachable Sets (GRS), against a state-of-the-art sampling-based planner, ContactRRT~\cite{suh2025ctr}, to address three key questions:

\begin{enumerate}
    \item Does our global optimization strategy lead to higher-quality plans than SOTA?
    \item Does our object-centric approach enable sub-minute query times? 
    \item Does the use of reachable sets produce kinematically and dynamically feasible plans across a range of queries?
\end{enumerate}

We find that our new planning paradigm excels along all three fronts—producing higher-quality plans, maintaining sub-minute query times, and reliably generating a plans for different queries. We also analyze which design choices account for these improvements upon SOTA.

Next, in Sec.~\ref{sec:ablation_experiments}, we validate that two of our add-on features further boost plan quality and success rates. We conclude that these additions to our core algorithm are crucial to elevating performance from good to excellent. 

Finally, in Sec.~\ref{sec:hardware_experiments}, we transfer our plans to hardware. Most plans are executed reliably, but some struggle at regrasps due to precision issues in $\pi$. We detail these challenges and suggest some potential fixes for future work.

\begin{figure*}[h!]
\centering

\begin{minipage}{0.32\textwidth}
    \centering
    \subfloat[Number MRS vs. task cost]{
        \includegraphics[width=\linewidth]{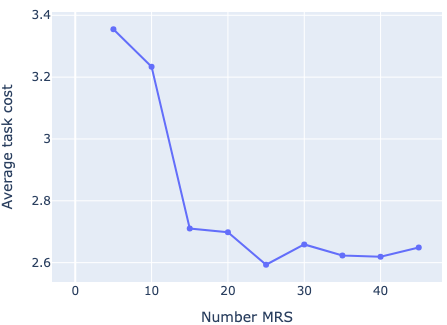}
    }
\end{minipage}
\hfill
\begin{minipage}{0.32\textwidth}
    \centering
    \subfloat[Planning success rate]{
        \includegraphics[width=\linewidth]{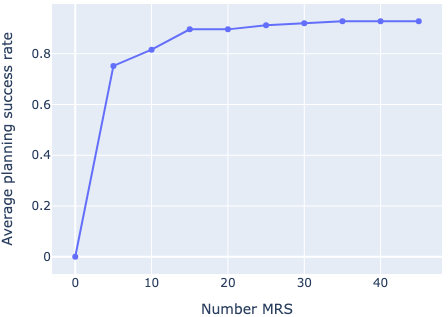}
    }
\end{minipage}
\hfill
\begin{minipage}{0.32\textwidth}
    \centering
    \subfloat[Offline time]{
        \includegraphics[width=\linewidth]{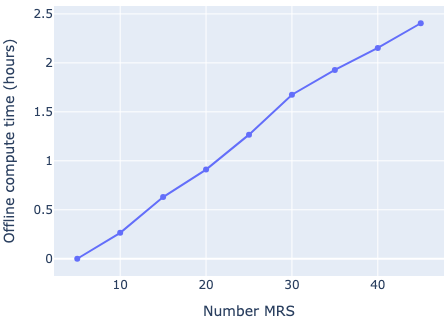}
    }
\end{minipage}

\caption{
Performance and compute time versus the number of MRS sets in the graph. 
Performance increases almost monotonically as more sets are added, providing a simple knob to trade off offline compute for higher-quality plans.
}
\label{fig:n_mrs_tuning}
\end{figure*}

\subsection{Experimental Setup}\label{sec:planning_experiments:exprimental_setup}

\subsubsection{System and Task} Our system and task are specified in Sec.~\ref{sec:task_specification}. We take goal tolerance $r = 0.1$ (\Cref{eq:prelim:goal_def}) and use the task cost in \eqref{eqn:our_task_cost}.

\subsubsection{Hardware and Implementation Details} The experiments are run on a desktop with an Intel(R) Core(TM) i9-9980XE CPU (18 cores, 36 threads) and 125 GB of RAM. The code is implemented in Python, with parallelization from Ray, optimization programs from Drake, GCS implementation from Drake, CQDC-MPC from~\cite{suh2025ctr}, and RRT-Connect from~\cite{kuffner2000rrt}. We simulate the controlled system by forward propagating the dynamics \eqref{eq:prelim:dynamics} under the action of an open-loop controller executing the action sequence.  

\subsubsection{Baselines}
For our baseline, we use ContactRRT~\cite{pang2023global}, a SOTA sampling-based planner for contact-rich manipulation. It is a variant of RRT~\cite{lavalle1998rapidly} with a contact-aware trajectory optimizer for the ``extend'' operation, which provides efficient exploration along the contact manifold. After the RRT step, it performs short-cutting to improve the plan. However, in the contact-rich setting, opportunities for short-cutting are limited because transitions cannot occur between arbitrary full configurations—their object configuration must match. Conveniently, ContactRRT uses our same contact-aware trajectory optimizer (CQDC-MPC) as a subroutine, so this allows us to directly compare our global planning strategies. While ContactRRT's global strategy focuses on finding feasible paths, GRS additionally seeks \textit{optimal} paths. We tune the hyperparameters for both GRS and ContactRRT and report results using the best-performing combinations (see \Cref{sec:appendix:hyperparam} for details).

\subsubsection{Metrics} We generate a test set of 250 queries, where $q^{o}_\text{start}, q^{o}_\text{goal}$ are sampled uniformly from the kinematic object workspace (Fig.~\ref{fig:system_illustration}). The following metrics are computed over the whole test set. To understand plan quality, we propose three metrics: 
\begin{enumerate}
    \item Task Cost: choice of cost function that defines the task, in the form of \eqref{eq:prelim:cost}. See Sec.~\ref{sec:prelim:prob_formulation} for examples.
    \item Object Travel Distance Ratio: from~\cite{cheng2023enhancing}, this is the total object travel distance divided by the straight-line start-to-goal distance:
    \begin{align}
        \rho_{travel} = \frac{\sum_{t=1}^{T-1} \norm{q^{o, *}_{t} - q^{o, *}_{t+1}}}{\norm{q^{o}_{start} - q^{o}_{goal}}}.
    \end{align}
    \item Robot Contact Change Ratio: from~\cite{cheng2023enhancing}, this is the total number of regrasps divded by the configuration path length:
    \begin{align}
        \rho_{contact} = \frac{\# \; \text{of regrasps}}{\sum_{t=1}^{T-1} \norm{q^{*}_{t} - q^{*}_{t+1}}}.
    \end{align}
\end{enumerate}

\noindent We also compute:
\begin{enumerate}[start=4] 
    \item Offline Time: time to build the graph offline,
    \item Query Time: time of plan inference online, 
    \item Planning Success Rate: the percent of queries for which the planner finds a kinematically and dynamically feasible solution.  
\end{enumerate}

\subsection{Results and Discussion}\label{sec:planning_experiments:results_discussion}

\begin{figure*}[tbp]
    \centering
    \begin{minipage}{0.48\textwidth}
        \centering
        \subfloat[]{
            \includegraphics[width=\linewidth]{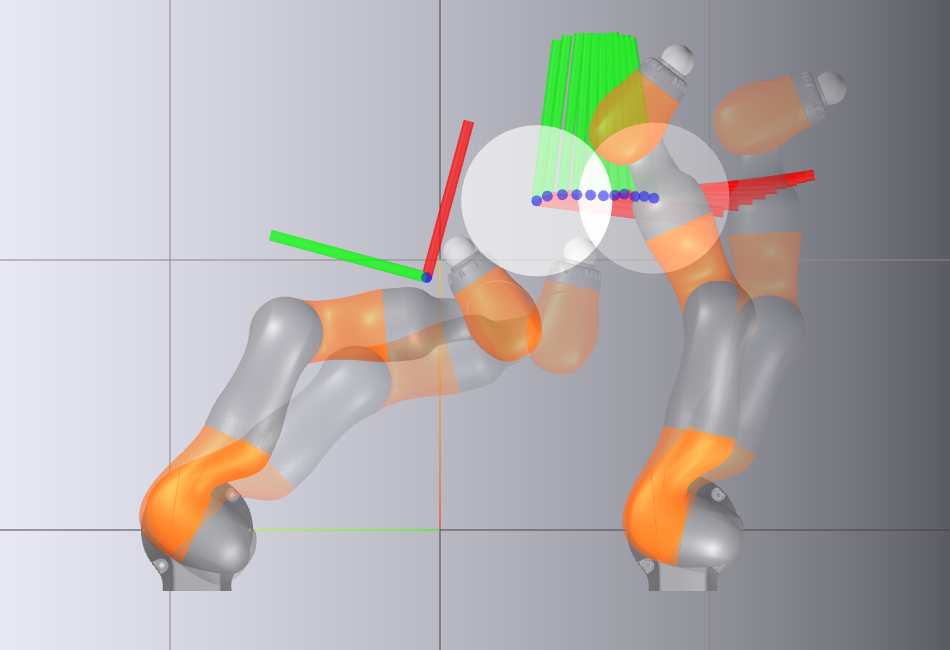}
        } \\
        \subfloat[]{
            \includegraphics[width=\linewidth]{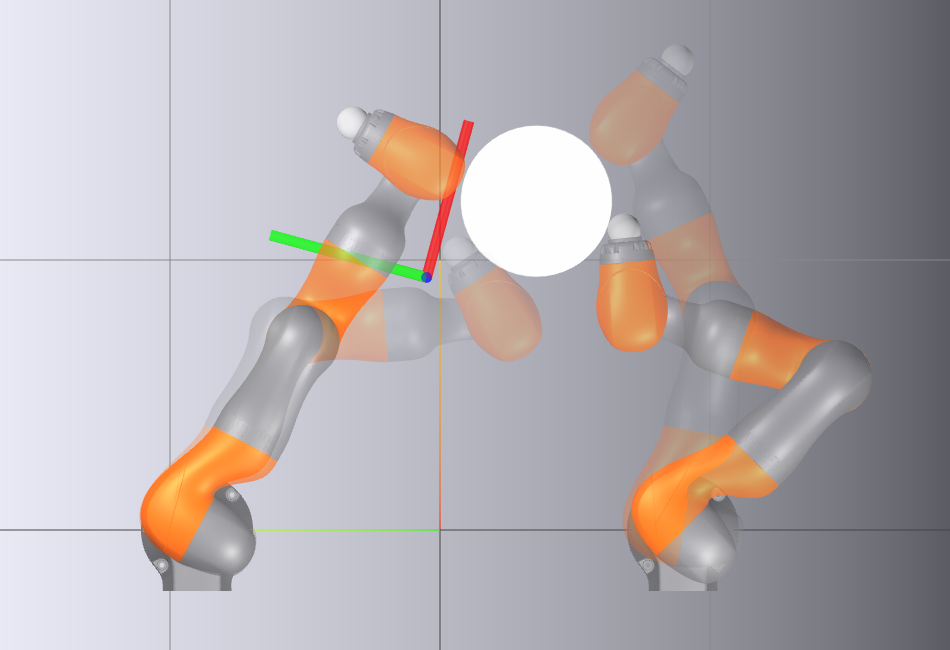}
        } \\
        \subfloat[]{
            \includegraphics[width=\linewidth]{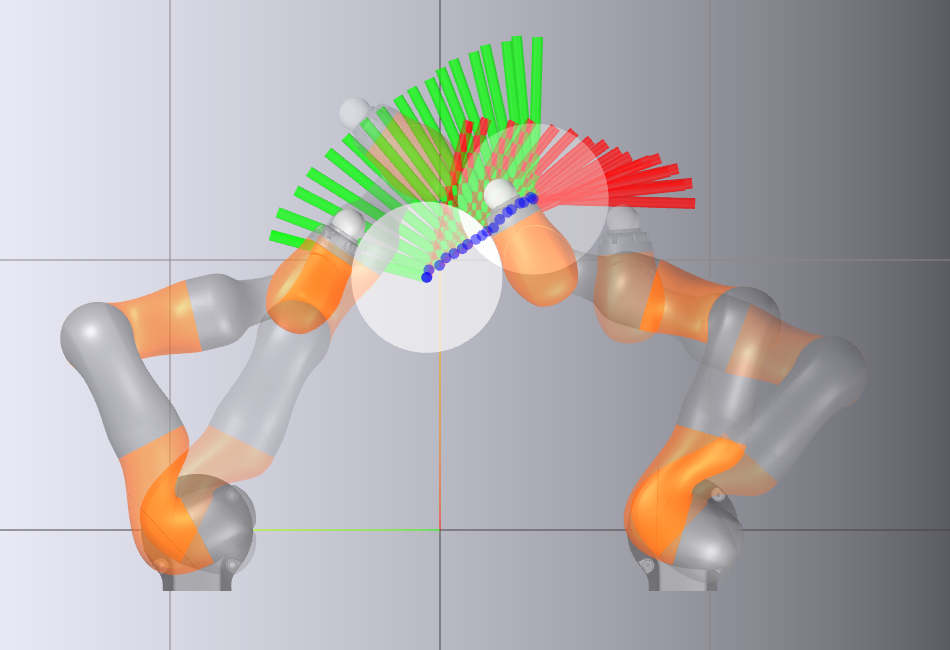}
        }
    \end{minipage}
    \hfill
    \begin{minipage}{0.48\textwidth}
        \centering
        \subfloat[]{
            \includegraphics[width=0.67\linewidth]{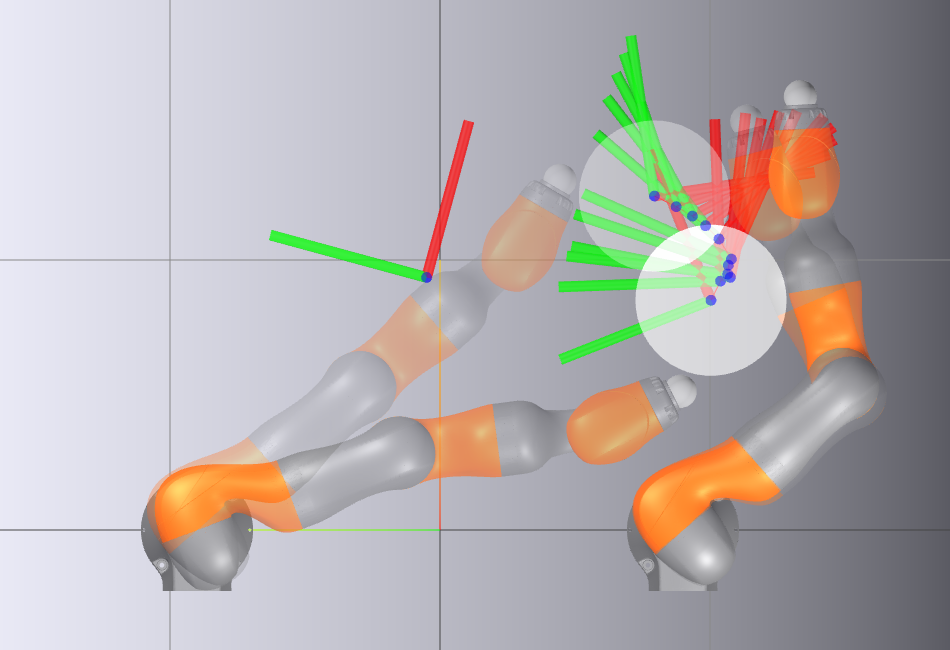}
        } \\
        \subfloat[]{
            \includegraphics[width=0.67\linewidth]{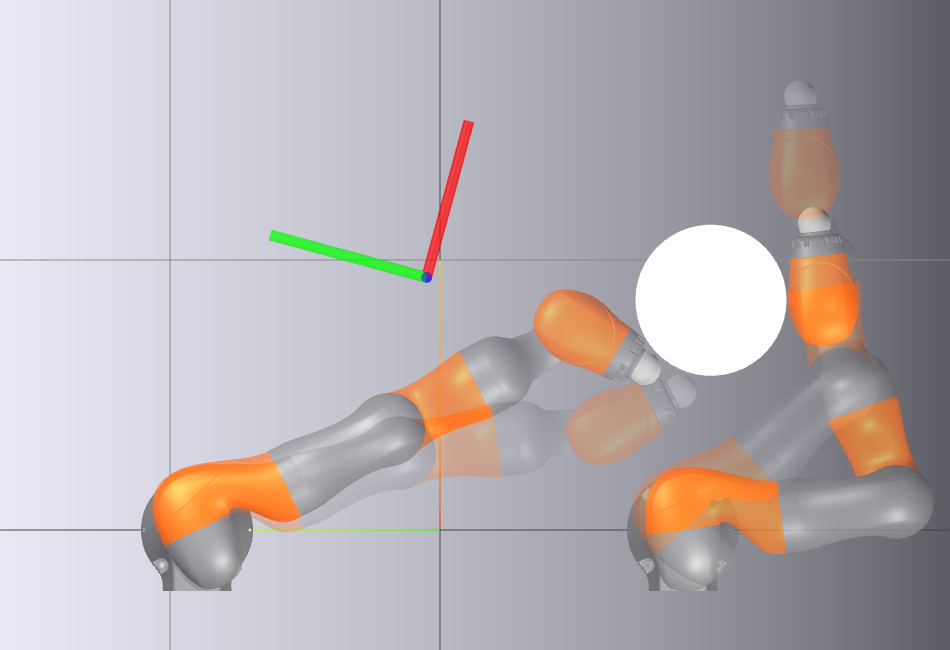}
        } \\
        \subfloat[]{
            \includegraphics[width=0.67\linewidth]{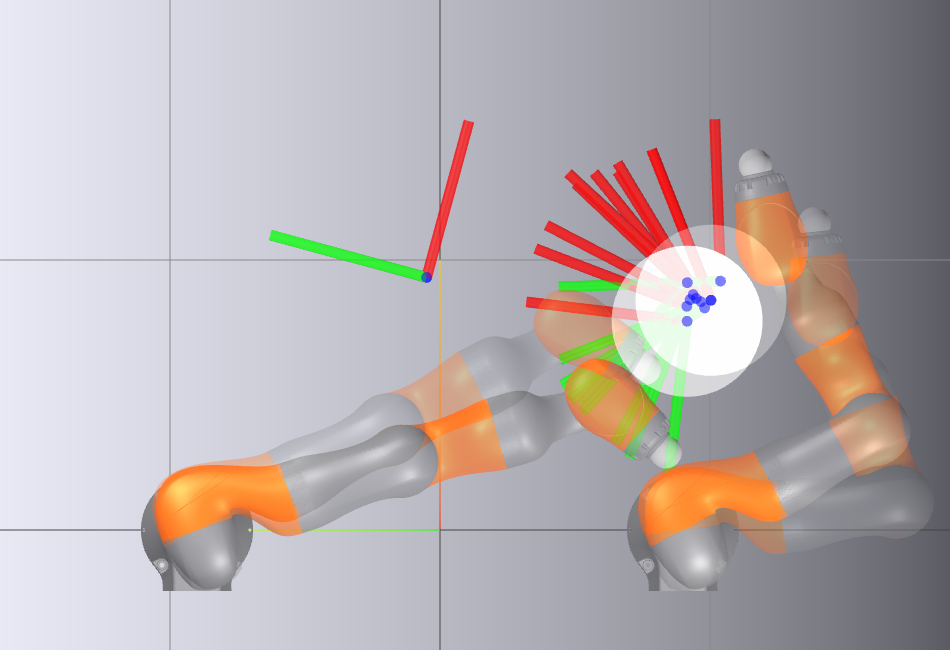}
        } \\
        \subfloat[]{
            \includegraphics[width=0.67\linewidth]{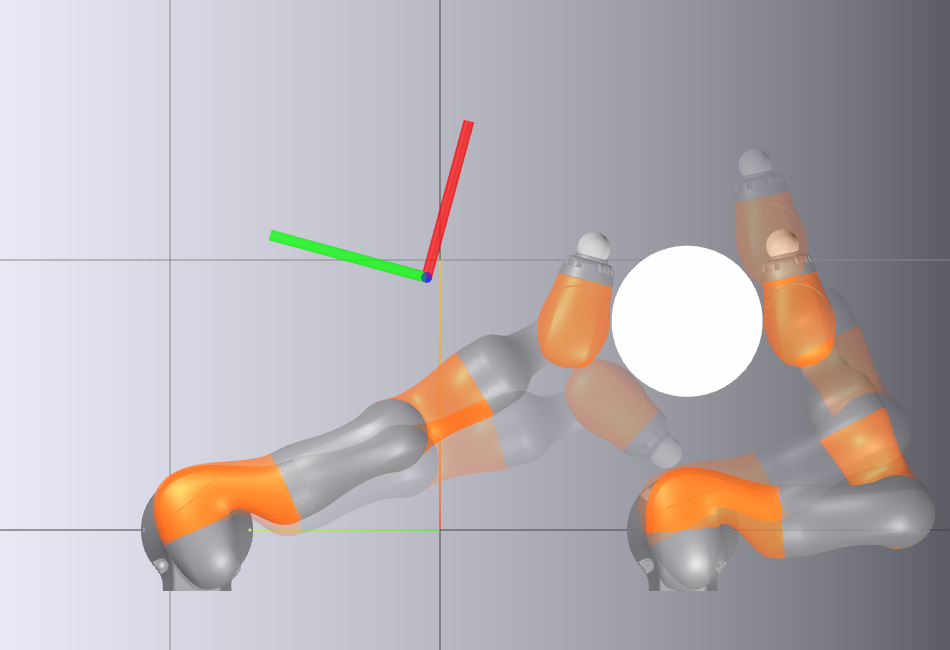}
        } \\
        \subfloat[]{
            \includegraphics[width=0.67\linewidth]{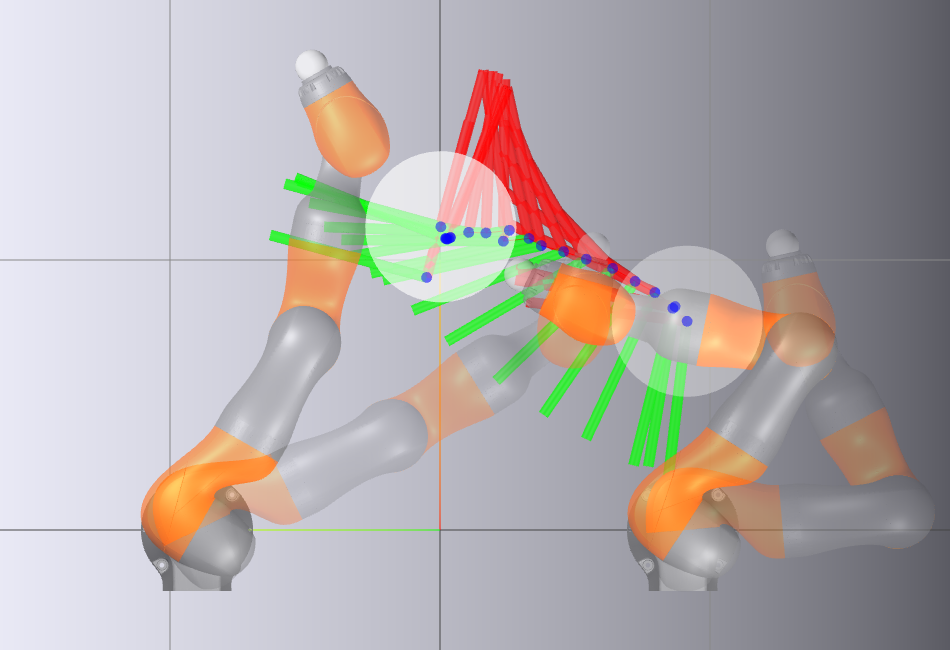}
        }
    \end{minipage}

    \caption{
    Illustration of planner rollouts for GRS (left column) and Contact-RRT (right column) on a given query. The images in each column show alternating transfer and transit segments, with the target object orientation overlaid. Both planners produce dexterous, contact-rich behavior (e.g., rolling the bucket along the arm), but GRS uses fewer regrasps by selecting grasps that maximize the length of transfer motions.
    }
    \label{fig:keyframes_query_157}
\end{figure*}

\begin{figure*}[tbp]
    \centering
    \begin{minipage}{0.30\textwidth}
        \centering
        \subfloat[]{
            \includegraphics[width=\linewidth]{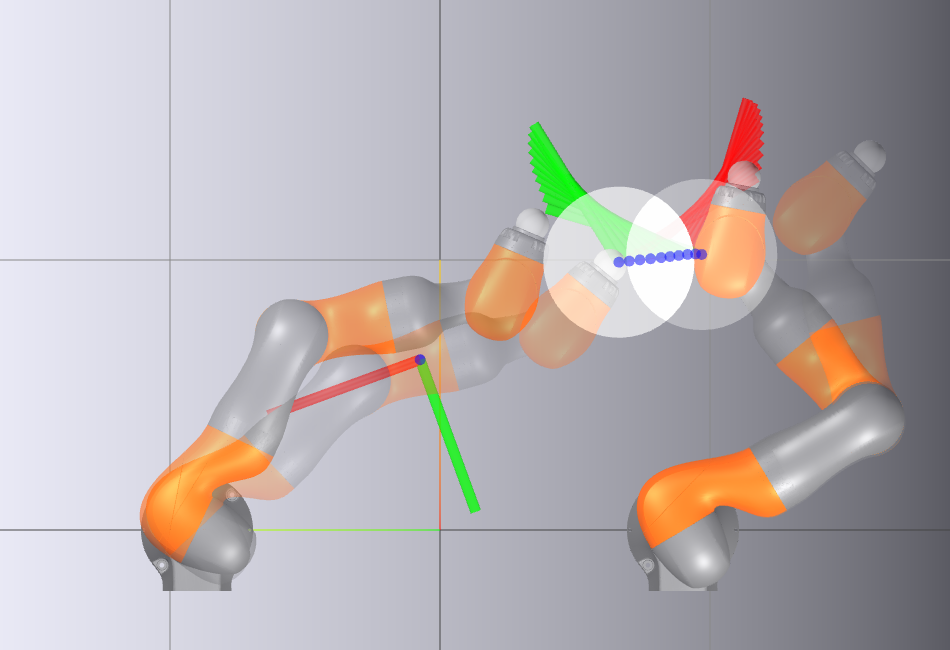}
        } \\
        \subfloat[]{
            \includegraphics[width=\linewidth]{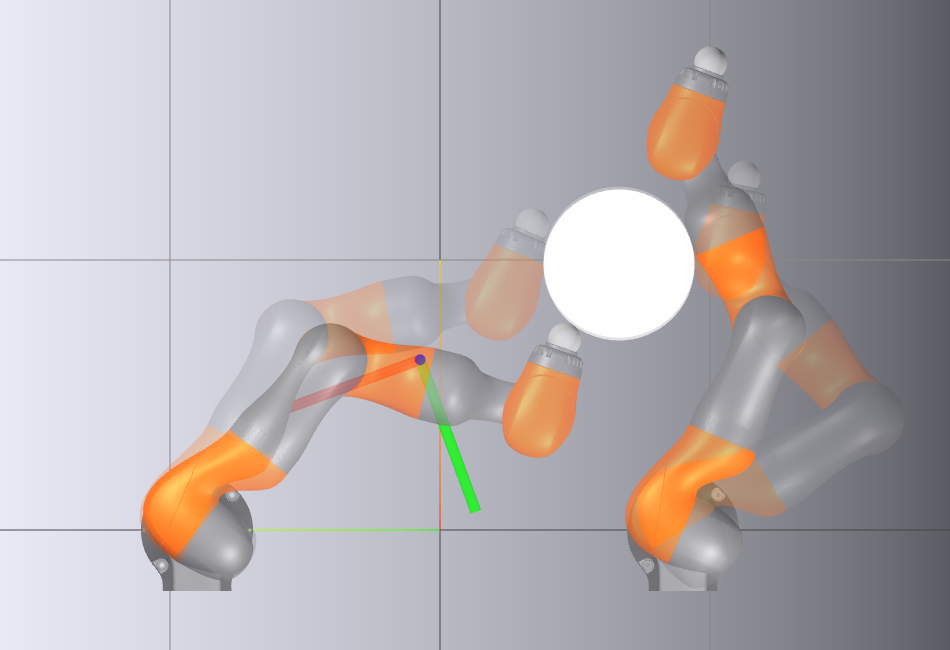}
        } \\
        \subfloat[]{
            \includegraphics[width=\linewidth]{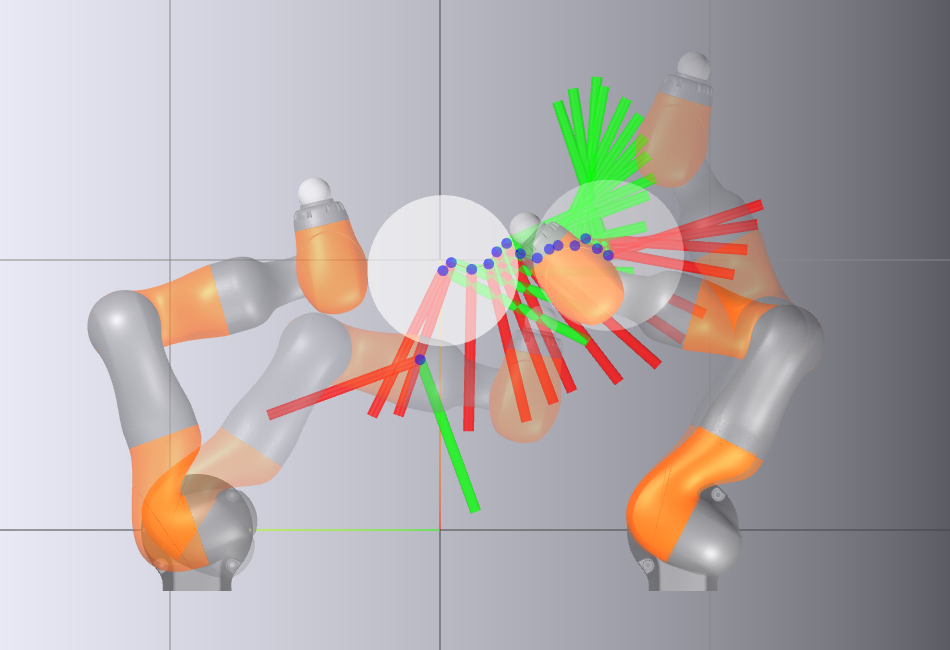}
        } \\
        \subfloat[]{
            \includegraphics[width=\linewidth]{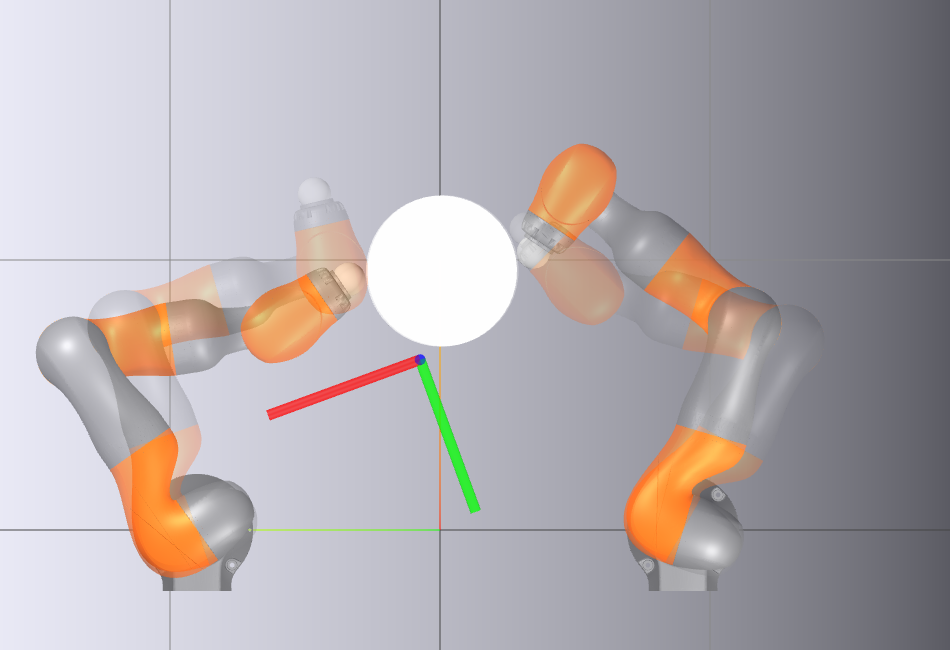}
        } \\
        \subfloat[]{
            \includegraphics[width=\linewidth]{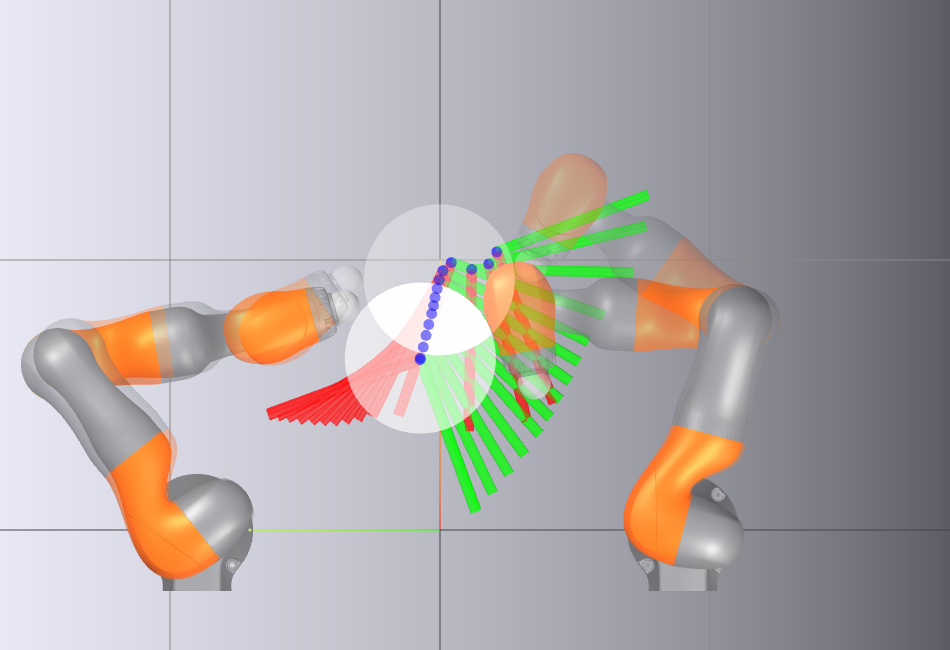}
        }
    \end{minipage}
    \hspace{0.3cm}
    \vrule width 1pt height 322pt
    \hspace{0.3cm}
    \begin{minipage}{0.30\textwidth}
        \centering
        \subfloat[]{
            \includegraphics[width=\linewidth]{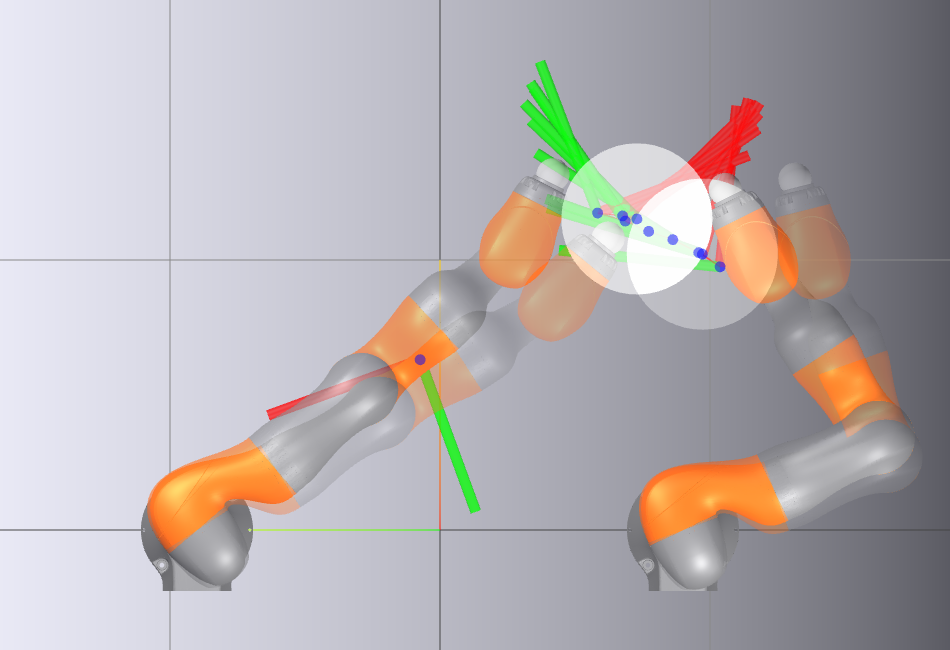}
        } \\
        \subfloat[]{
            \includegraphics[width=\linewidth]{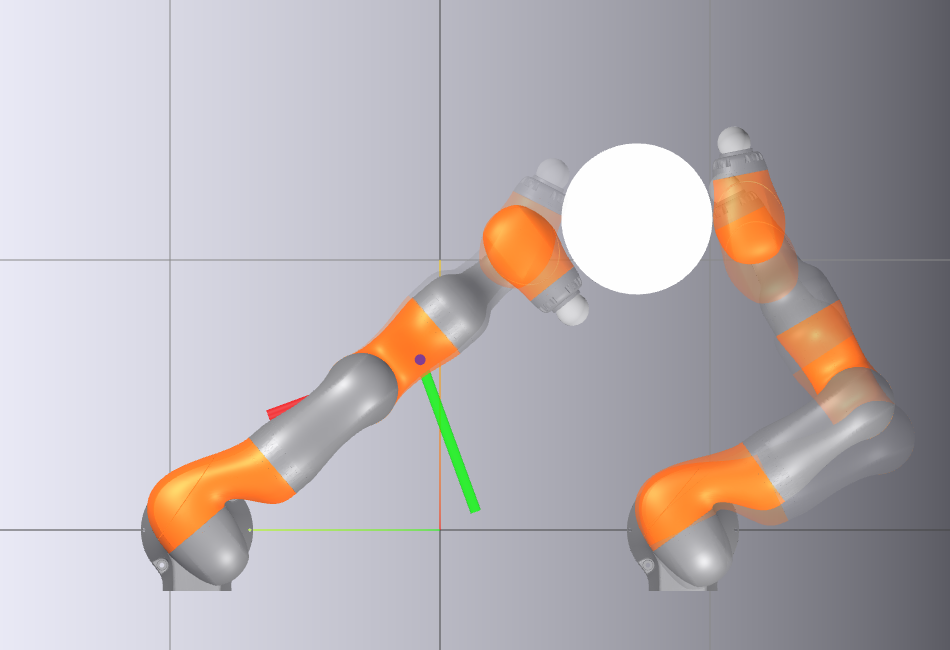}
        } \\
        \subfloat[]{
            \includegraphics[width=\linewidth]{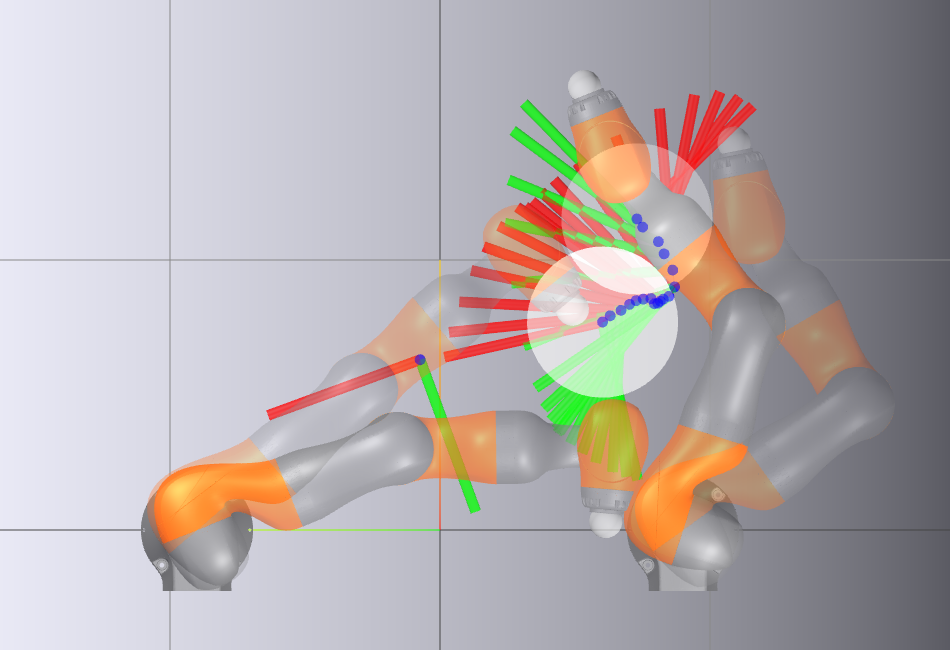}
        } \\
        \subfloat[]{
            \includegraphics[width=\linewidth]{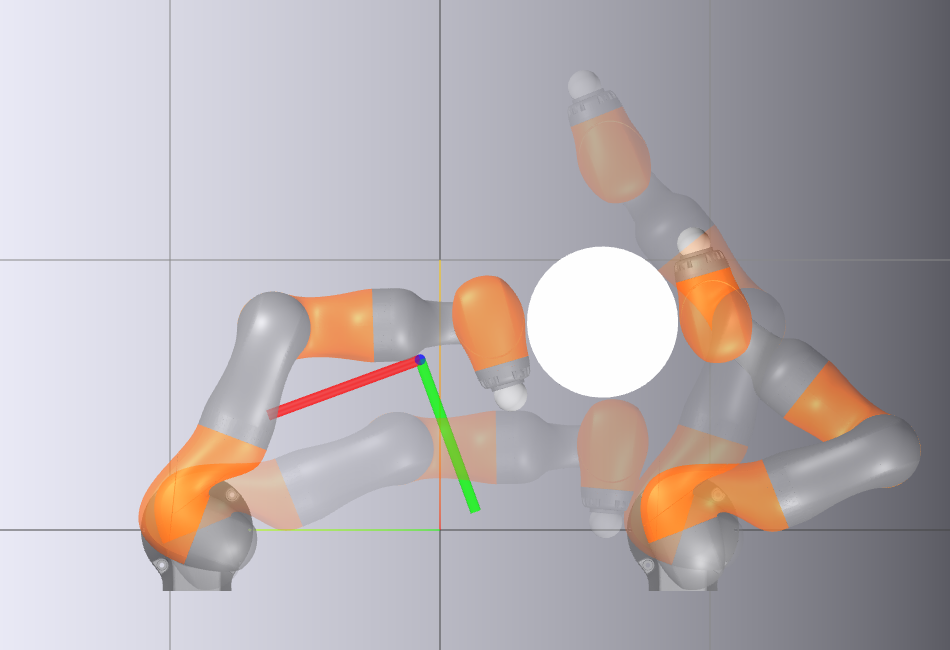}
        } \\
        \subfloat[]{
            \includegraphics[width=\linewidth]{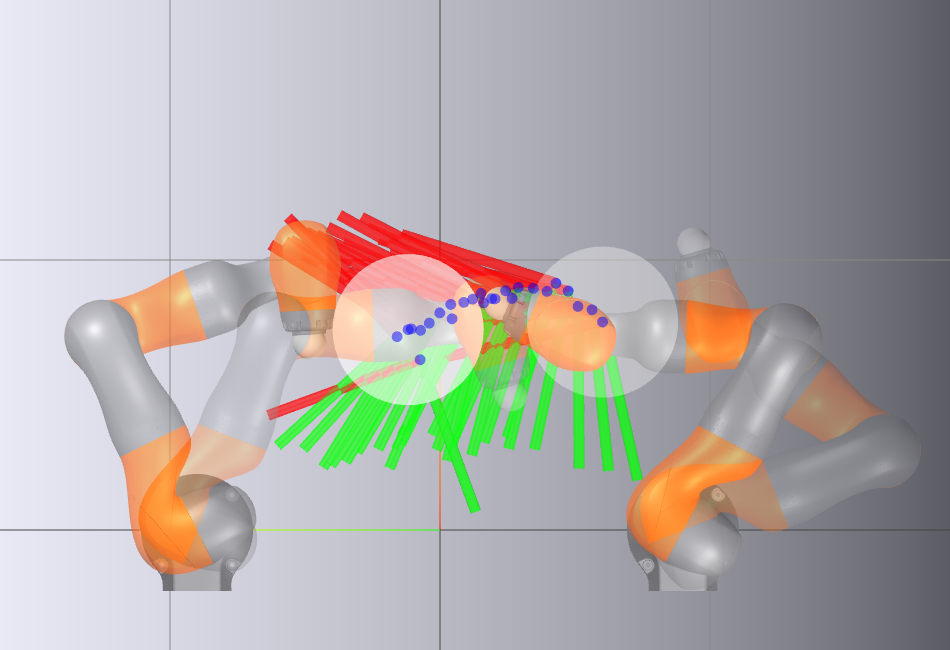}
        }
    \end{minipage}
    \hfill
    \begin{minipage}{0.30\textwidth}
        \centering
        \subfloat[]{
            \includegraphics[width=\linewidth]{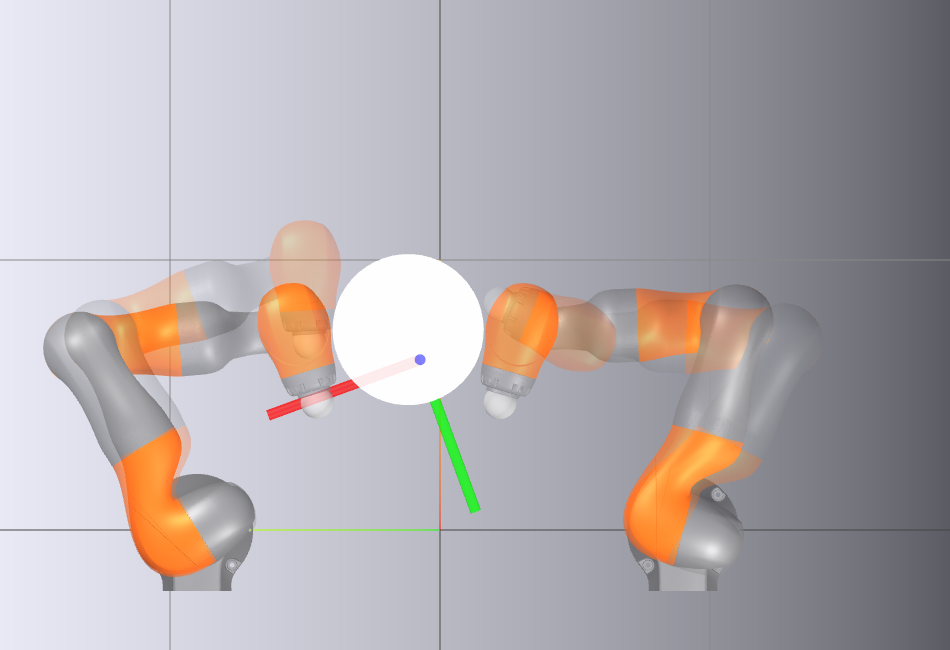}
        } \\
        \subfloat[]{
            \includegraphics[width=\linewidth]{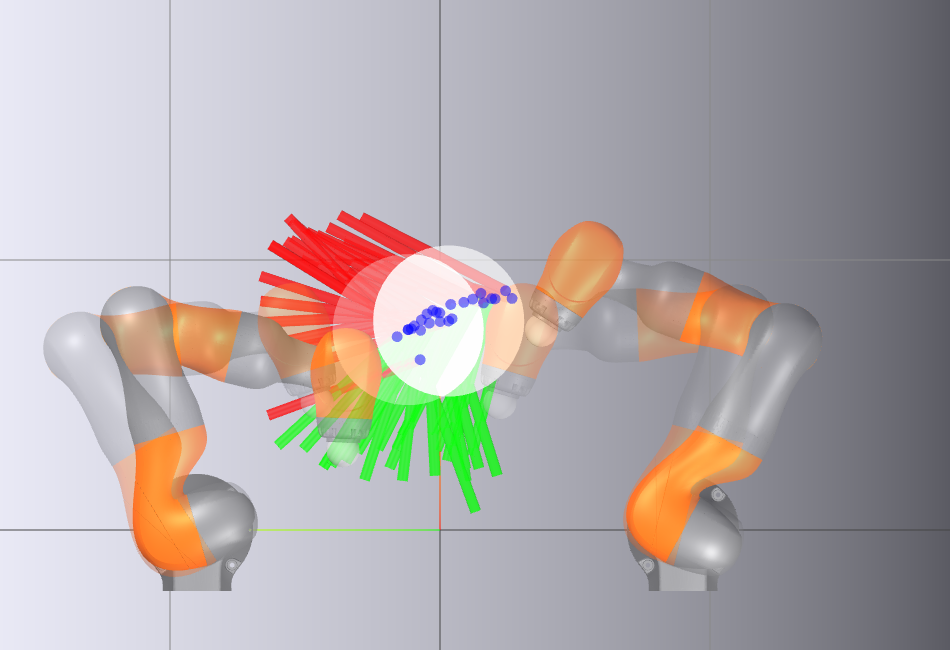}
        } 
    \end{minipage}
    \caption{
Planner rollouts for GRS (left column) and Contact-RRT (middle and right columns) on another given query. The images in each column show alternating transfer and transit segments. This query requires a large rotation (nearly 
$180^\circ$). Contact-RRT requires one more regrasp than GRS, and the quality of its transfer motions also differs: while GRS moves the object along straight, goal-directed paths, Contact-RRT often produces curved or jagged paths that may momentarily move the object away from the goal due to its sampling-based exploration.}
\label{fig:keyframes_query_187}
\end{figure*}

\begin{table*}[h]
\centering
\caption{
Our method, GRS, compared to baseline ContactRRT on path quality, planning success rate, and query time over a test dataset of 250 randomly sampled queries.} 
\begin{tabular}{|c|c|c|c|c|c|}
\hline
& Task Cost & Object Travel Distance Ratio & Robot Contact Change Ratio & Planning Success Rate & Query Time (sec) \\ \hline
ContactRRT & $6.62 \pm 5.41$ & $3.70 \pm 3.23$ & $0.10 \pm 0.094$ & $82.4\%$ & $62.04 \pm 130.33$ \\ 
GRS (Ours) & $2.59 \pm 1.47$ & $1.19 \pm 0.36$ & $0.075 \pm 0.11$ & $91.2\%$ & $40.34 \pm 24.73$ \\ \hline   
\end{tabular}
\label{tab:vs_baseline}
\end{table*}

\begin{figure}[tbp]
    \centering
    \includegraphics[width=\linewidth]{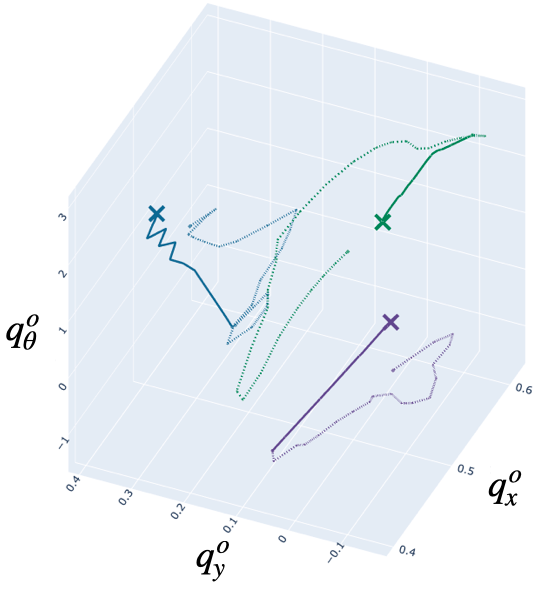}
    \caption{Three example plans (green, blue, purple) for GRS (solid line) and ContactRRT (dashed line). The goals are marked by ``x''. Note that GRS's plans tend to be concise, whereas ContactRRT's tend to be circuitous.
    }  
    \label{fig:chosen_gcsr_vs_rrt_se2}
\end{figure}

\subsubsection{Path Quality}\label{sec:planning_experiments:quality}
We first evaluate the quality of our planned paths. Our object-space plans are effectively optimal with respect to the fitted task cost, up to the choice of set cover and continuous path parameterization. As shown in \eqref{eqn:experiments:gcs_mip_gap}, the object plans computed using GCS are within 1\% of the optimal MIP solutions in cost, while requiring only about 5\% of MIP’s runtime.

\begin{align}
    \delta_{opt} = \frac{\mathcal{C}_{\mathrm{GCS}} - \mathcal{C}_{\mathrm{MIP}}}{\mathcal{C}_{\mathrm{MIP}}} = 0.0145 \pm 0.0607 \label{eqn:experiments:gcs_mip_gap}
\end{align}

After translating to full configuration space, the resulting plans remain \textit{approximately optimal}, achieving substantially lower costs than the SOTA baseline. As summarized in Table~\ref{tab:vs_baseline}, GRS reduces the average task cost by nearly 61\%, where the cost metric \eqref{eqn:our_task_cost} measures path length in actuated configuration space, encouraging short robot trajectories and penalizing actuation effort. This strong correspondence between optimal object-space plans and high-quality full-space plans further validates our object-centric formulation.

GRS also achieves significantly lower auxiliary costs that correlate with task performance. It yields a nearly 68\% lower object travel distance ratio, meaning that not only are the paths in actuated space more succinct, but the object paths are too. As seen in Fig.~\ref{fig:chosen_gcsr_vs_rrt_se2}, ContactRRT often produces “switchback” or meandering trajectories, while GRS produces short, direct paths. These kinds of paths reflect ContactRRT's random growth strategy. Evidently, ContactRRT's short-cutting offers limited improvement too, as it cannot address this underlying exploration bias. Additionally, GRS exhibits a 25\% reduction in the robot contact-change ratio, since under our task metric, regrasps are costly and naturally minimized.

Comparing the two planners visually highlights their qualitative differences (Fig.~\ref{fig:keyframes_query_157},~\ref{fig:keyframes_query_187}). GRS's superior path quality can also be explained by its ability to decompose a large reorientation into just a few screw motions. The GCS optimization selects grasps that maximize maneuverability along these intended screws. In essence, GRS leverages global optimization to generate concise, natural motions, often resembling those a human might use. We also observe the merits of the contact-rich manipulation paradigm: it has a substantial advantage over end-effector manipulation where rotation is involved. Large rotations can be accomplished by rolling along manipulator surfaces; end-effector manipulator is limited to small rotations.

\subsubsection{Planning Success Rate}\label{sec:planning_experiments:success_rate} We evaluate planning success rate across a diverse test set of 250 queries, where $(q^{o}_{\text{start}}, q^{o}_{\text{goal}})$ are uniformly sampled from the kinematic object workspace (Fig.~\ref{fig:system_illustration}). Planning success rate is defined as the percentage of queries for which the algorithm produces a kinematically and dynamically feasible plan. Possible sources of failure for GRS are discussed in Sec.~\ref{sec:method:online:sources_of_failure}.

GRS achieves a nearly $11\%$ higher success rate than ContactRRT. At first, this may seem surprising: the ContactRRT timeout was set high enough for all queries to complete, and sampling-based planners are typically good at exploring configuration space and finding solutions. However, a closer look at the failure cases reveals key differences in the two methods.

It turns out nearly all of ContactRRT’s failures stem from transit failures. Like GRS, it assumes that arbitrary transits are feasible: i.e. that a collision-free path exists between any two grasps.\footnote{In $\mathrm{SE}(2)$, this is the only requirement for transit. Typically, maintaining static equilibrium of the object is also required. However, here, the object can simply be left resting on the plane.} However,  prior to transit, sometimes the manipulator gets trapped behind the object or stuck at a singularity (Fig.~\ref{fig:transit_failure}). This is likely due to ContactRRT's step size -  a large step is needed to explore the space efficiently, but such a step size can also send the arm to extreme configurations from which it is hard to transit.

On the other hand, GRS has high planning success rates by design.
Hierarchical planners often fail when their high-level plans cannot be translated into feasible low-level plans. This occurs if the high-level planner is unaware of feasibility constraints - but this is not the case for GRS. Its high-level object space planning uses reachable sets, which embed information about manipulator feasibility.  

Specifically, each transfer segment within an object plan corresponds to a motion inside an MRS, which is feasible by definition (Lemma~\ref{lemma:prelim:mutual_reachability}). Each transit segment is assumed feasible, but unlike in ContactRRT, this assumption generally holds because GRS avoids problematic configurations prior to transit. This is because all configurations belong to an MRS, which by definition excludes singular or trapped states: each configuration must be able to reach the seed configuration, ensuring recoverability (\Cref{dfn:prelim:mrs}). 

\subsubsection{Offline and Online Runtimes}\label{sec:planning_expexperiments:runtimes} 
During offline graph construction, we use a set cover approximation constant of $\alpha = 0.98$ (Sec.~\ref{sec:method:offline:covering}), resulting in a graph of 25 sets (Fig.~\ref{fig:set_cover}).  
Computing these discretized MRS requires about 800k total $(q, u, q^{+})$ simulation steps, which are generated while attempting to reach grid cells under the local trajectory optimizer $\pi$. This corresponds to 22.25 hours of simulated time, or 1.72 hours of wall-clock time on our machine. Because we can attempt to reach each grid cell independently, MRS generation is highly parallelizable, resulting in short wall-clock times.

Plan quality and success rate can be monotonically improved by increasing the number of MRS in the graph, at the cost of longer offline compute times (Fig.~\ref{fig:n_mrs_tuning}). This highlights another advantage of our approach: it provides a natural way to trade off performance and computation by adjusting the granularity of the discrete decision space. Other popular methods lack this mechanism~\cite{cheng2022contact, hauser2010multi}.

Online, GRS computes most plans in under a minute, while ContactRRT takes 55\% longer. This is as expected, since GRS is multi-query and has shifted much of the computational burden to the offline stage. Further, to breakdown the GRS runtime: running GCS takes $0.71$ seconds on average and translating the object path with $\pi$, $\psi$ takes $39.63$ seconds. 

\section{Ablation Experiments}\label{sec:ablation_experiments}
In this section, we validate that our two additional features contribute meaningfully to planning performance. We aim to answer the following questions:
\begin{enumerate}
    \item Does path sampling improve success rate?
    \item Do fitted GCS costs improve plan quality?
\end{enumerate}
We conduct an ablation study comparing planner performance with and without these features (Table~\ref{tab:ablation}) and find that both yield noticeable performance gains.  

\subsubsection{Path Sampling}\label{sec:ablation:path_sampling}
In the variant with path sampling (\textit{``All features''} in Table~\ref{tab:ablation}), we sample up to 100 distinct paths, averaging $9.53 \pm 13.40$ samples per query.   
In the variant without path sampling (\textit{``Without path sampling''} in Table~\ref{tab:ablation}), GCS returns only the single lowest-cost path. We find that including this feature improves the success rate by $7.5\%$.  

This result highlights the benefit of path sampling: it addresses multiple sources of failure simultaneously, including transfer failures caused by MRS mis-approximation and transit failures due to violated assumptions. Because these errors are difficult to eliminate completely, path sampling serves as a practical mitigation strategy. However, it does double the online runtime, so it can optionally be omitted.

\subsubsection{Fitted Surrogate GCS Vertex and Edge Costs}\label{sec:ablation:fitted_costs}
In the run with fitted GCS costs (\textit{``All features''} in Table~\ref{tab:ablation}), we generate costs using the procedure in Sec.~\ref{sec:method:offline:cost_fitting}.  
In the run without fitted costs (\textit{``Without fitted GCS costs''} in Table~\ref{tab:ablation}), we use simple heuristic costs: a constant edge cost of 10 for regrasps, and an $\ell_2$-norm cost for move-object edges.  
We find that using fitted costs reduces overall plan cost by $5.79\%$, confirming that learned cost models serve as better proxies for true task costs than heuristics.  

Overall, these results show that path sampling and fitted GCS costs improve success rate and average plan cost by roughly $8\%$ and $6\%$, respectively. Given the competitiveness of the SOTA baseline, these gains represent a meaningful improvement in planner capability.

\begin{table}[h]
\centering
\caption{
Ablation study to determine efficacy of two features: path sampling and fitted GCS costs. Statistics computed over a testset of 250 randomly sampled queries.
}
\begin{tabular}{|c|c|c|c|}
\hline
& Task Cost &  \makecell{Planning \\Success Rate} & \makecell{Query Time \\ (seconds)} \\ \hline
All features & $2.59 \pm 1.47$ & $91.2\%$ & $40.34 \pm 24.73$ \\
No path sampling & $2.97 \pm 2.09$ & $84.8\%$ & $19.11 \pm 13.70$ \\
No fitted GCS costs & $2.74 \pm 1.89$ & $87.6\%$ & $27.32 \pm 28.50$ \\ \hline
\end{tabular}
\label{tab:ablation}
\end{table}

\section{Hardware Experiments}\label{sec:hardware_experiments}

\begin{figure*}[t!]
\centering
\begin{minipage}{0.29\textwidth}
    \centering
    \subfloat{
        \includegraphics[width=\linewidth]{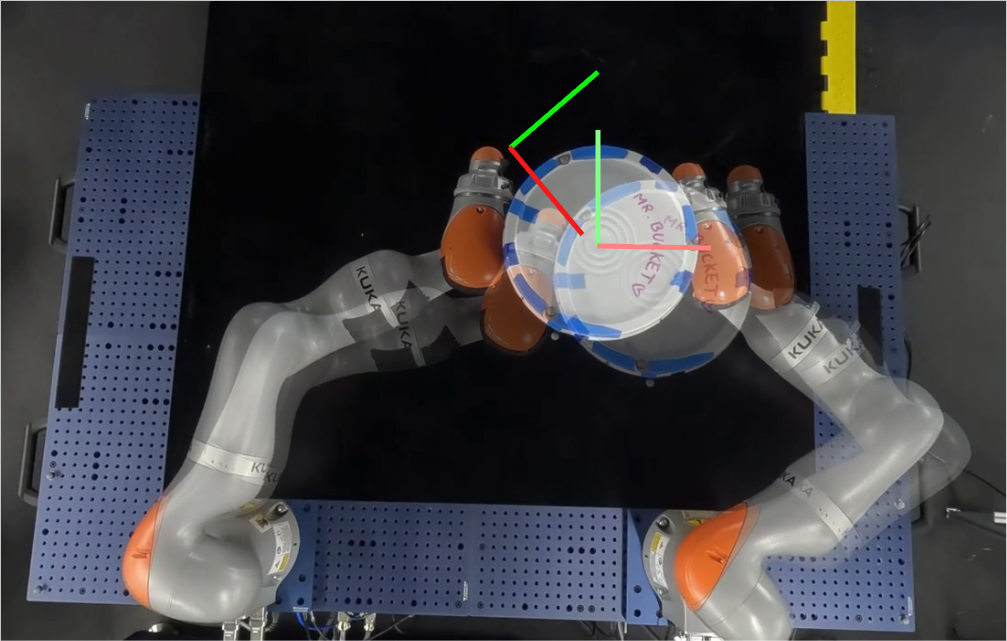}
    }
\end{minipage}
\hfill
\begin{minipage}{0.29\textwidth}
    \centering
    \subfloat{
        \includegraphics[width=\linewidth]{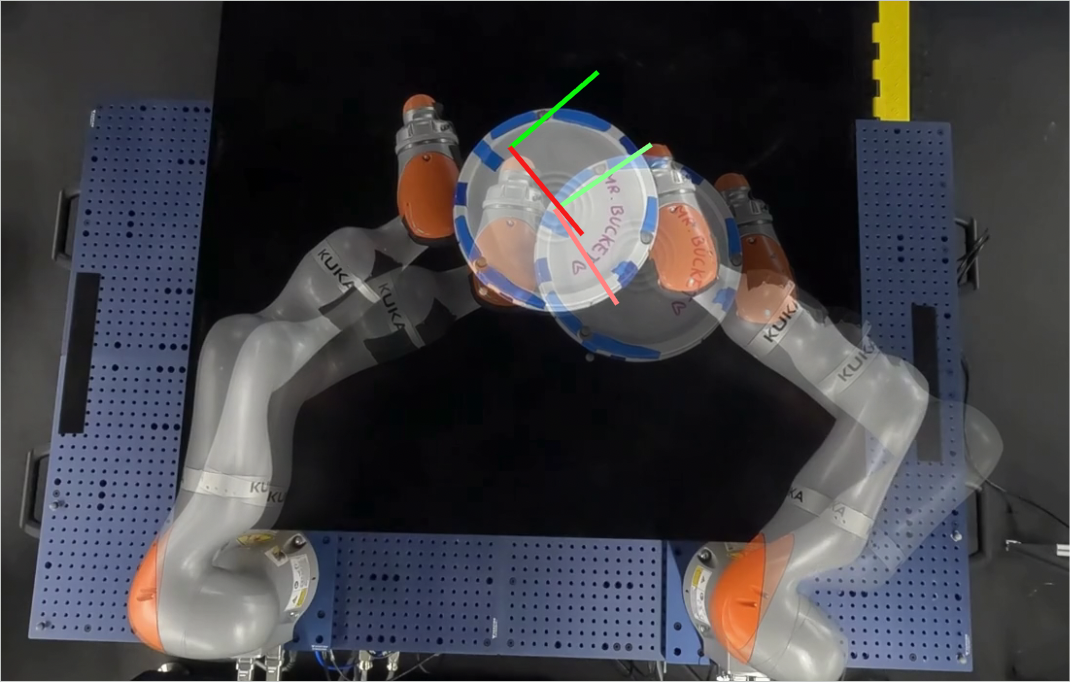}
    }
\end{minipage}
\hfill
\begin{minipage}{0.29\textwidth}
    \centering
    \subfloat{
        \includegraphics[width=\linewidth]{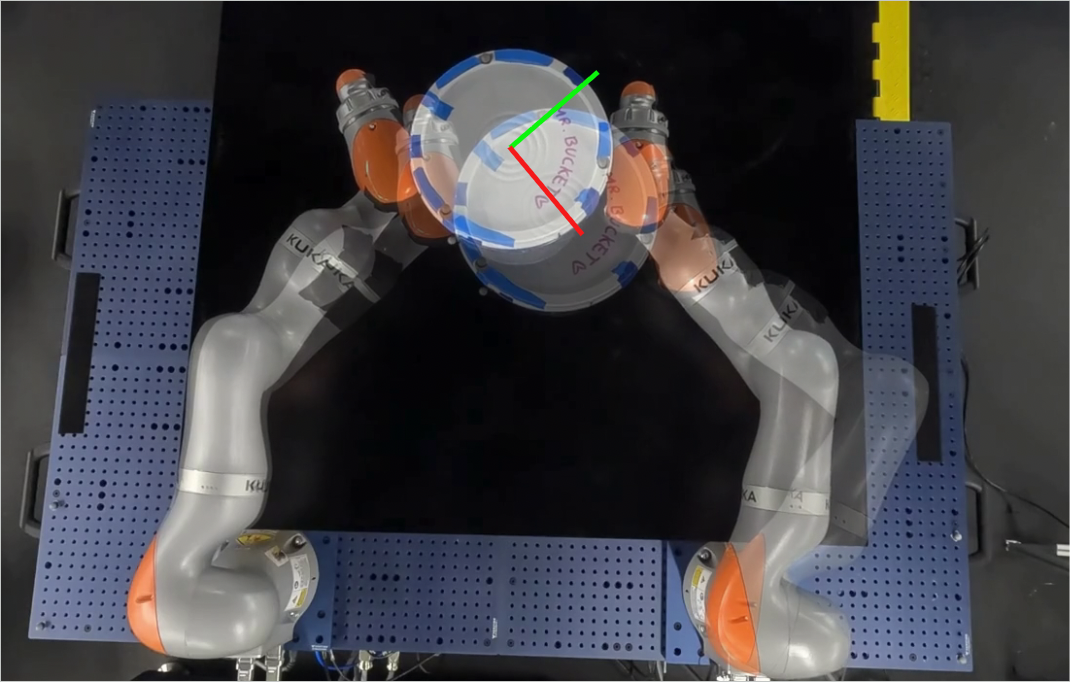}
    }
\end{minipage}
\caption{
Keyframes from a GRS plan executed on hardware, with the target orientation overlaid on all frames. This plan contains only one transfer segment. Observe the contact-rich behavior, especially in the final frame, where the full surface of the “hand” is used to guide the object into precise alignment.
}
\label{fig:keyframes_hardware_q1}
\end{figure*}

Finally, we evaluate our approach on hardware. After computing the manipulator input sequence, we execute the plan open-loop. Rollouts without regrasps execute reliably (Fig.~\ref{fig:keyframes_hardware_q1}), indicating that the CQDC dynamics model transfers well to hardware in our task regime (SE(2), slow motions).

In contrast, rollouts that include regrasps succeed only intermittently. Two factors contribute to regrasp failure: numerical imprecision when computing the new grasp, and state drift accumulated during open-loop execution. To explain these further, first suppose the planned state immediately before a regrasp is $ q $, and the next planned MRS is $ \mathcal{R}' $.

Recall that the new grasp is computed using the local planner $ \pi $. Specifically, to ensure the grasp lies within $ \mathcal{R}' $, we use $ \pi $ to reach from the seed of $ \mathcal{R}' $ to the object state $ q^o $, which is assumed to remain stationary during the regrasp. However, due to limited numerical precision, $ \pi $ may only reach the target up to an accuracy of $ \epsilon_\pi $, yielding an object mismatch
\[
d_{\mathrm{SE}(2)}(q^{\prime,o}, q^o) \le \epsilon_\pi .
\]
As a result, the manipulator may attempt to regrasp the object at a slightly offset pose. While small mismatches can sometimes be tolerated, larger deviations often cause the regrasp to fail.

The second source of failure is drift introduced during open-loop execution. The regrasp target is computed from the planned state $ q $, while execution error the true state $ \tilde{q} $ to deviate, increasing misalignment at regrasp time. A straightforward remedy is to replan online rather than execute open-loop. In particular, the target grasp can be recomputed as $ \mathbf{proj}^{-1}_{\mathcal{Q}^{o}}(\tilde{q}^o, \mathcal{R}') $ instead of $ \mathbf{proj}^{-1}_{\mathcal{Q}^{o}}(q^o, \mathcal{R}') $. We leave this extension to future work.

\section{Conclusion}
In this paper, we introduced an \emph{intermediate-level} discrete decision space, mutual reachable sets, situated between traditional low-level contact modes and high-level motion primitives.  
This abstraction is fine-grained enough to handle a wide range of queries, yet coarse enough to shrink the decision space dramatically, enabling optimal planning in object space.

Our planner (GRS) outperforms a SOTA sampling-based planner on a challenging, representative contact-rich task. It achieves a $61\%$ reduction in task cost (producing more natural, concise plans with full-arm contact), a $91\%$ planning success rate across 250 queries, and sub-minute query times. With our new planning paradigm, it is now possible to compute such contact-rich plans for tasks of real-world complexity.

We view this as a first step toward a broader family of reachable-set abstractions for planning. For example, we can imagine extensions to $\mathrm{SE}(3)$, which will require new methods to handle Riemannian space. 

\appendix
\section{Appendix}\label{sec:appendix}

\subsection{Contact-Aware Trajectory Optimizer Details}\label{sec:appendix:ctr}

In this section, we describe the CQDC trajectory optimization program and how we apply it in an MPC-fashion as a local planner. 
Recall that CQDC assumes a quasidynamic, discrete-time model~\cite{pang2023global}.

\begin{align}
q_{+} = f(q, u) \label{eq:prelim:dynamics} 
\end{align}

\noindent When used for control, this model is locally smoothed and then linearized:

\begin{align}
\hat{q}_{+} = \matr{A_\kappa} \partial q + \matr{B_\kappa}\partial u + f_{\kappa}(\Bar{q}, \Bar{u}) \\
\matr{A_\kappa} \defeq \partial f_{\kappa}/ \partial q(\Bar{q}, \Bar{u}), \matr{B_\kappa} \defeq \partial f_{\kappa} / \partial u(\Bar{q}, \Bar{u}) \\
\partial q \defeq q - \Bar{q}, \partial u \defeq u - \Bar{u}
\end{align}
where $f_{\kappa}$ is the smoothed dynamics, implemented as a log-barrier relaxation of the forward convex program with parameter $\kappa$~\cite{boyd2004convex}.

\noindent The linearized model is constrained to a trust region~\cite{suh2025ctr}. Then, we form a convex trajectory optimization program where we provide a starting configuration $q_0$, a state reference trajectory $\bar{q}_{0:T}$, an input reference trajectory $\bar{u}_{0:T-1}$, and a goal configuration $q_{\text{goal}}$. The initial guess for the first timestep is provided by a collection of heuristics; see \cite{suh2025ctr} for more details. This program solves for a control sequence $\partial u_{0:T-1}^{*}$ that reaches toward the goal and follows the reference. 
\begin{align}
    & \text{CQDCTrajOpt}(q_{0}, \bar{q}_{0:T}, \bar{u}_{0:T-1}, q_{\text{goal}}) = \partial u_{0:T-1}^{*}, \text{where} \\
    & \min_{\delta q_{0:T},\delta u_{0:T-1}} \quad  \|q_\text{goal} - q_T\|^2_\mathbf{Q} + c(q_{0:T})  \label{eq:linear_trajopt:cost}\\
    & \text{s.t.} \quad \delta q_{t+1}= \mathbf{A}_{\kappa, t} \delta q_t + \mathbf{B}_{\kappa, t} \delta u_t, \; t = 0\dots T-1, \label{eq:linear_trajopt:linear_dynamics_constraint} \\
    & \quad \quad (\delta q_t, \delta u_t)\in \tilde{\mathcal{S}}_\mathbf{\Sigma, \kappa}(\bar{q}_t,\bar{u}_t), \; t = 0\dots T-1,\label{eq:linear_trajopt:ctr}\\
    & \quad \quad q_t = \bar{q}_t + \delta q_t, \; t = 0\dots T, \\
    & \quad \quad u_t = \bar{u}_t + \delta u_t, \; t = 0\dots T-1,\\
    & \quad \quad |u_t - u_{t-1}| \leq \eta, \; t = 1\dots T-1,\\
    & \quad \quad \delta q_0 = 0, \label{eq:linear_trajopt:initial_condition}
\end{align}
where $\tilde{\mathcal{S}}_\mathbf{\Sigma,\kappa}$ in \eqref{eq:linear_trajopt:ctr} is a trust region. 

We solve this program in an MPC fashion (CQDC-MPC)~\Cref{alg:mpc} to generate local plans. To improve goal-reaching reliability, we run each goal-reaching query with a set of different state reference trajectories. These trajectories all start at $q_0$ and end with $q_{\text{goal}}$ but may have different midpoints. We find that makes goal-reaching succeed more often. 




\begin{algorithm}
\caption{MPC}\label{alg:mpc}
\textbf{Input:} Initial state $q_0$, goal state $q_{\text{goal}}$\;
\textbf{Output:} Lists of visited states $L_q$, applied inputs $L_u$\;
$L_q \leftarrow [q_0]$, $L_u \leftarrow$ \texttt{list()}\;
\For {$t = 0$ to $H - 1$} {
    \eIf{$t == 0$} { \label{alg:mpc:begin_if}
        $\bar{u}_{0:T-1}\leftarrow$ Apply initial guess heuristics; \label{alg:mpc:first_initialization}
    } {
        $\bar{u}_{0:T-1}\leftarrow$ Initialize from the previous $u_{0:T-1}^\star$\; \label{alg:mpc:later_initialization}
    }
    $\bar{q}_{0:T-1}\leftarrow$ Compute state reference from $q_t$, $q_{\text{goal}}$; \label{alg:mpc:q_ref}\;
    $u_{0:T-1}^\star \leftarrow \mathbf{CQDCTrajOpt}(q_t, \bar{q}_{0:T}, \bar{u}_{0:T-1}, q_{\text{goal}}$)\; \label{alg:mpc:trajopt}
    $q_{t+1} = f(q_t, u_0^\star)$\; \label{alg:mpc:dynamics_rollout}
    $L_q$.\texttt{append}($q_{t+1}$), $L_u$.\texttt{append}($u_0^\star$)\;
}
\Return $\; L_q, \; L_u$
\end{algorithm}

 
\subsection{Collision-Free Motion Planner Details} \label{sec:appendix:collision_free}

For collision free motion planning, we use use RRT-Connect \cite{kuffner2000rrt} to generate a feasible collision-free path, and then refine with trajectory optimization. In this section, we elaborate on the trajectory optimization portion of the collision free planner.

Suppose RRT-Connect returns the sequence of robot configurations which describe a collision free path for the robot that brings the system from $q_0$ to $q_T$ without making contact with the object. To minimize the length of the collision free path, we solve the following nonconvex optimization program:

\begin{subequations}
\begin{align}\label{eq:appendix:trajopt}
    q^{a,\star}_{0:T} = \argmin_{q^a_{0:T}} &\sum_{t=0}^{T-1} \| q^a_{t+1} - q^a_t \|^2_{Q} \\
    \text{s.t.} \;\;
    & \phi_j(q_t) \ge \epsilon, \quad \forall j \; \forall t \label{eq:appendix:trajopt:collision} \\
    & q^a_\text{lb} \leq q^a_t \leq q^a_\text{ub}, \quad \forall t
\end{align}
\end{subequations}
where $Q$ is a cost matrix that penalizes robot movements, $\phi_j$ denotes the signed distance function for the $j$-th collision pair, $\epsilon$ denotes the minimum acceptable distance between the robot and the object, and $q^a_\text{lb}$ and $q^a_\text{ub}$ denote the lower and upper joint limits for the robot.

We solve this nonconvex program in an iterative manner, using sequential quadratic programming (SQP). In each iteration, constraint \Cref{eq:appendix:trajopt:collision} is linearized around the solution to the previous iteration. We use RRT-Connect's output $\bar{q}^a_{0:T}$ as an initial guess solution and return $q^{a,\star}_{0:T}$.

\subsection{Proof of Lemma 1}\label{sec:appendix:lemma1}

\setcounter{lemma}{0}
\begin{lemma}
Suppose $q^o_1$ and $q^o_2$ are elements of the MRS $\mathcal{R}^o$. Then $q^o_1$ is reachable from $\mathbf{proj}_{\mathcal{Q}^{o}}^{-1}(q^o_2, \mathcal{R}^{o})$, and $q^o_2$ is reachable from $\mathbf{proj}_{\mathcal{Q}^{o}}^{-1}(q^o_1)$.
\end{lemma}



In the following proof, we implicitly assume that if $\pi$ can find a path between $q_1$ and $q^o_\text{seed})$, and also between $q_\text{seed}$ and $q^o_2$, it can find a path between $q_1$ and $q^o_2$. This is actually true by design of our $\pi$, since for any query, we give it a reference trajectory with $q_{\text{seed}}$ as a midpoint. 

\begin{proof}
By symmetry, it suffices to show that $q^o_2$ is reachable from $q_1^o$.
Let $q_1 = \mathbf{proj}_{\mathcal{Q}^{o}}^{-1}(q^o_1, \mathcal{R}^o)$ and $q_2 = \mathbf{proj}_{\mathcal{Q}^{o}}^{-1}(q^o_2, \mathcal{R}^o)$. Since $q_1 \in \mathcal{R}^{-}$, $\pi$ can find a path between $q_1$ and $q_\text{seed})$. Further, since $q_2 \in \mathcal{R}^{-}$, $\pi$ can find a path between $q_\text{seed}$ and $q_2$. Hence, a path exists from $q_1$ to $q_2$ through $q_{\text{seed}}$ and $\pi$ can find it. 
\end{proof}

\subsection{Grasp Generation for $q^{o}_{\text{seed}}$ on Bimanual Kuka} \label{sec:appendix:qseed}

We compute simple antipodal grasps with the wrist joints touching opposite sides of the object. We define a mathematical program for computing such a grasp. 

First, for $i=1,2$, our two desired contact points, we define a function $f^{\text{a}}_i: \mathcal{Q}^a \to \mathbb{R}^3$ that maps the robot configuration to a point on the robot's surface in the world frame, and a corresponding function $f^{\text{o}}_i: \mathcal{Q}^o \to \mathbb{R}^3$ that maps the object configuration to a point on the object's surface in the world frame. Given an object configuration $q^o$, we obtain a grasping robot configuration $q^{a,\star}_\text{grasp}$ by solving the following inverse kinematics problem:
\begin{subequations}\label{eqn:appendix:contact_sampler}
\begin{align}
    q^{a,\star}_{\text{grasp}} = \argmin_{q^a_\text{grasp}} &\sum_{i=1}^N \frac{1}{2}\norm{f^a_i(q^a_\text{grasp}) - f^o_i(q^o)}^2_2 \\
    \text{s.t.} \;\; & \phi_j((q^a_\text{grasp}, q^o)) \ge 0, \quad \forall j \label{eqn:appendix:contact_sampler:nonpenetration}\\ 
                    & q^a_\text{grasp} \in \mathcal{Q}^a \label{eqn:appendix:contact_sampler:joint_limit}
\end{align}
\end{subequations}
where \ref{eqn:appendix:contact_sampler:nonpenetration} enforces nonpenetration constraints between the robot and the object and \ref{eqn:appendix:contact_sampler:joint_limit} enforces joint limit constraints for the robot. The resulting seed configuration $q_\text{seed} = (q^{a,\star}_\text{grasp}, q^o)$ is returned as output. We solve this nonconvex program using \textsc{Snopt} \cite{snopt}, provided with \textsc{Drake} \cite{drake}.

In our system, we set $N=2$. We define $f^a_1$ such that it maps to a point on the left arm's wrist joint, and $f^o_1$ such that it maps to a point on the left side of the object. We define $f^a_2$ and $f^o_2$ in a similar manner for the right arm/right side of the object. Additionally, $f^o_1$ and $f^o_2$ are constructed to map to points on opposite sides of the object.

\subsection{Convex Sets in $\mathrm{SE}(2)$} \label{sec:appendix:se2_convexity}

We adopt the GCS viewpoint of Cohn et al. \cite{cohn2025noneuclidean}. The object workspace $\mathcal{Q}^o \subseteq \mathrm{SE}(2)$ is regarded as a Riemannian manifold $(\mathcal{Q}^o,g)$ with the product metric introduced in \Cref{sec:prelim:prob_formulation}, whose geodesic distance is $d_{\mathrm{SE}(2)}$ in \Cref{eqn:dse2}. On $(\mathcal{Q}^o,g)$ we use the notion of geodesically convex ($g$-convex) sets from \cite[Sec.~3.2]{cohn2025noneuclidean}: a set is $g$-convex if, for any two points in the set, the unique minimizing geodesic between them lies entirely in the set.

Cohn et al.\ show that $\mathbb{R}^2 \times S^1$ with this product metric is a flat configuration manifold and that, on such flat product manifolds, GGCS reduces to an ordinary Euclidean GCS: under any chart that is a local isometry, a subset is $g$-convex whenever its image is Euclidean convex and its diameter in each factor is smaller than the convexity radius \cite[Assumption 2 and Theorem 4]{cohn2025noneuclidean}. In our case we use the global chart $\psi(q^o) = (q^o_x,q^o_y,q^o_\theta) \in \mathbb{R}^3$, and the convex-approximated mutual reachable sets $\hat{\mathcal{R}}^o_{\Delta}$ are computed as convex polytopes in these coordinates. We additionally enforce that the angular projection of each $\hat{\mathcal{R}}^o_{\Delta}$ has width strictly less than $\pi$, so that minimizing geodesics between points in the set do not wrap around $S^1$. By the results of \cite{cohn2025noneuclidean}, each $\hat{\mathcal{R}}^o_{\Delta}$ is therefore $g$-convex in $(\mathcal{Q}^o,g)$, and the GCS problem we solve in these coordinates is the Euclidean reduction of the corresponding GGCS problem on $\mathcal{Q}^o$.



\subsection{Probabilistic Completeness of \Cref{alg:cover}} \label{sec:appendix:cover}

In this section, we prove that for any $\alpha$, the procedure described in \Cref{alg:cover} can produce an $\alpha$-approximate cover. We make the following four assumptions: 1) we assume $\mathcal{Q}^o$ to be measurable and of finite measure; 2) we assume that in \Cref{alg:mrs}, \texttt{IrisZo} always includes the cell in which $q^o_\text{seed}$ belongs; 3) we assume \texttt{IrisZo} does not include any regions not in $\mathcal{Q}^o$; and 4) we assume that \texttt{GenerateGrasp} is always successful.

\begin{proof}

Fix $\alpha$. For discretization resolution $\delta$, define $\mathcal{Q}^o_{\text{in}, \delta}$ to be the finite union of all cells of which are wholly contained within $\mathcal{Q}^o$. From analysis, we know that for every $\epsilon > 0$, there exists a $\delta > 0$ such that $\mathrm{vol}(\mathcal{Q}^o \setminus \mathcal{Q}^o_{\text{in},\delta}) \le \epsilon$. Thus, there exists a $\delta > 0$ such that $\mathrm{vol}(\mathcal{Q}^o_{\text{in},\delta}) \ge \alpha \cdot \mathrm{vol}(\mathcal{Q}^o)$.

Pick $\delta$ such that $\mathrm{vol}(\mathcal{Q}^o_{\text{in},\delta}) \ge \alpha \cdot \mathrm{vol}(\mathcal{Q}^o)$. Note that $\mathcal{Q}^o_{\text{in}, \delta}$ is a union of \emph{finitely} many cells. Because each MRS $\hat{\mathcal{R}}_\delta^o$ contains the cell in which its seed object configuration belongs (by assumption), we know that finitely many MRS are sufficient to include every cell in $\mathcal{Q}^o_{\text{in},\delta}$. Because each $q^o_\text{seed}$ is sampled from the uncovered portion of $\mathcal{Q}^o$, and because a grasp can always be synthesized around $q^o_\text{seed}$, we know that each iteration of the \textbf{while} loop (lines 3-6 of \Cref{alg:cover}) introduces a cell that was previously at least partially uncovered into $C$. Our third assumption guarantees that the union of all the MRS within $C$ do not include any regions outside of $\mathcal{Q}^o$.

Therefore, finitely many \textbf{while} loop iterations are sufficient for $C$ to include every cell in $\mathcal{Q}^o_{\text{in}, \delta}$. It follows that $\mathrm{vol}(\bigcup_{i=1}^{|C|}\hat{\mathcal{R}}^{o,i}_{\delta}) \ge \mathrm{vol}(\mathcal{Q}^o_{\text{in},\delta}) \ge \alpha \cdot \mathrm{vol}(Q^o)$, as desired.

\end{proof}

\subsection{Fitting Surrogate GCS Vertex and Edge Costs}\label{sec:appendix:cost}

In this section, we describe in detail how to fit surrogate edge and vertex costs using least-squares regression.

\subsubsection{Edge Costs (Transit Motions)}

Consider an edge $e_{ij}$ connecting sets $\mathcal{R}^i$ and $\mathcal{R}^j$ with seed configurations $q^i_{\text{seed}}$ and $q^j_{\text{seed}}$. This edge represents a transit motion that keeps the object fixed at some $q^o \in \mathcal{R}^i \cap \mathcal{R}^j$. Let
\[
q_0 = \mathbf{proj}^{-1}_{\mathcal{Q}^o,\pi,q^i_{\text{seed}}}(q^o), \quad
q_T = \mathbf{proj}^{-1}_{\mathcal{Q}^o,\pi,q^j_{\text{seed}}}(q^o)
\]
denote the corresponding full configurations.

Running the transit planner $\psi$ from $q_0$ to $q_T$ yields a trajectory $q_{0:T}$ with cost
$\ell_{e_{ij}}(q^o, q^o) = c(q_{0:T})$.
Since this cost is not available in closed-form and is not convex in $q^o$, we approximate it with a surrogate cost $\hat{\ell}_{e_{ij}}$.

We sample $K$ object configurations $\{q^o_k\}_{k=1}^K \subset \mathcal{R}^i \cap \mathcal{R}^j$, compute the corresponding trajectories using $\psi$, and record the costs $\{c(q_{0:T,k})\}$. Because $\psi$ (RRT-Connect) is stochastic, the resulting costs are multimodal. We therefore fit a constant surrogate, which minimizes the least-squares objective and preserves convexity and nonnegativity:
\begin{equation}
\label{eqn:appendix:final_edge_cost}
    \hat{\ell}_{e_{ij}}(q^o, q^o)
    =
    \frac{1}{K} \sum_{k=1}^{K} c(q_{0:T,k}) .
\end{equation}

\subsubsection{Vertex Costs (Transfer Motions)}

Consider a vertex $v_i$ corresponding to transfer motions within $\mathcal{R}^i$, with seed configuration $q^i_{\text{seed}}$. Let $q^o_0, q^o_T \in \mathcal{R}^i$ be the start and goal object configurations, and let
\[
q_0 = \mathbf{proj}^{-1}_{\mathcal{Q}^o,\pi,q^i_{\text{seed}}}(q^o_0)
\]
be the corresponding full configuration.

Executing the transfer planner $\pi$ from $q_0$ toward $q^o_T$ produces a trajectory $q_{0:T}$ with cost
$\ell_{v_i}(q^o_0, q^o_T) = c(q_{0:T})$.
As before, we approximate this black-box cost with a convex surrogate $\hat{\ell}_{v_i}$.

We sample $K$ pairs $(q^o_{0,k}, q^o_{T,k}) \in \mathcal{R}^i \times \mathcal{R}^i$, run $\pi$ to obtain trajectories and costs $c(q_{0:T,k})$, and fit a nonnegative convex quadratic. Let
\[
x_k =
\begin{bmatrix}
q^o_{0,k} \\
q^o_{T,k}
\end{bmatrix},
\qquad
y_k = c(q_{0:T,k}) .
\]
The surrogate has the form $\hat{y}_k = x_k^T A x_k + b^T x_k + c$, with parameters obtained by solving
\begin{subequations}
\label{eqn:appendix:vertex_costs}
\begin{align}
    \min_{A,b,c}\;\;
        & \sum_{k=1}^{K} \tfrac{1}{2}(\hat{y}_k - y_k)^2 \\
    \text{s.t.}\;\;
        & A \succeq 0, \\
        &
        \begin{bmatrix}
            A & \tfrac{1}{2} b^T \\
            \tfrac{1}{2} b & c
        \end{bmatrix}
        \succeq 0 ,
\end{align}
\end{subequations}
which enforce convexity and nonnegativity, respectively. The resulting vertex cost is
\begin{equation}
\label{eqn:appendix:final_vertex_cost}
    \hat{\ell}_{v_i}(q^o_0, q^o_T)
    =
    \begin{bmatrix}
        q^o_0 & q^o_T
    \end{bmatrix}
    A
    \begin{bmatrix}
        q^o_0 \\ q^o_T
    \end{bmatrix}
    +
    b^T
    \begin{bmatrix}
        q^o_0 \\ q^o_T
    \end{bmatrix}
    + c .
\end{equation}

\subsection{Additional Details on Planning Failures}
\begin{figure}[H]
\centering
\subfloat[Right arm trapped behind object.]{
    \includegraphics[width=0.45\columnwidth]{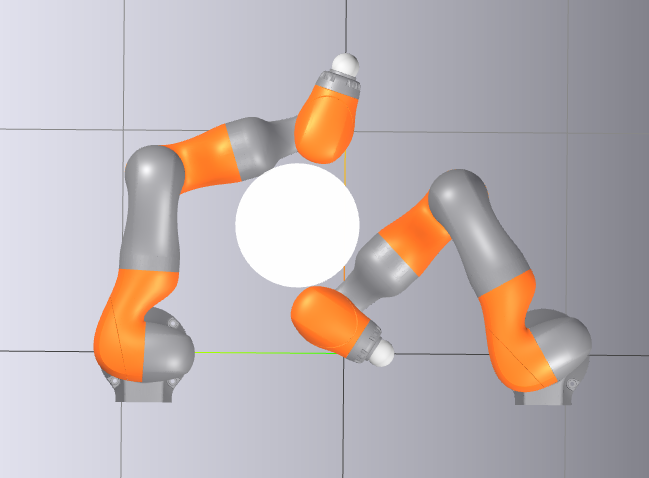}
}
\vspace{2pt}
\subfloat[Left arm at singularity.]{
    \includegraphics[width=0.45\columnwidth]{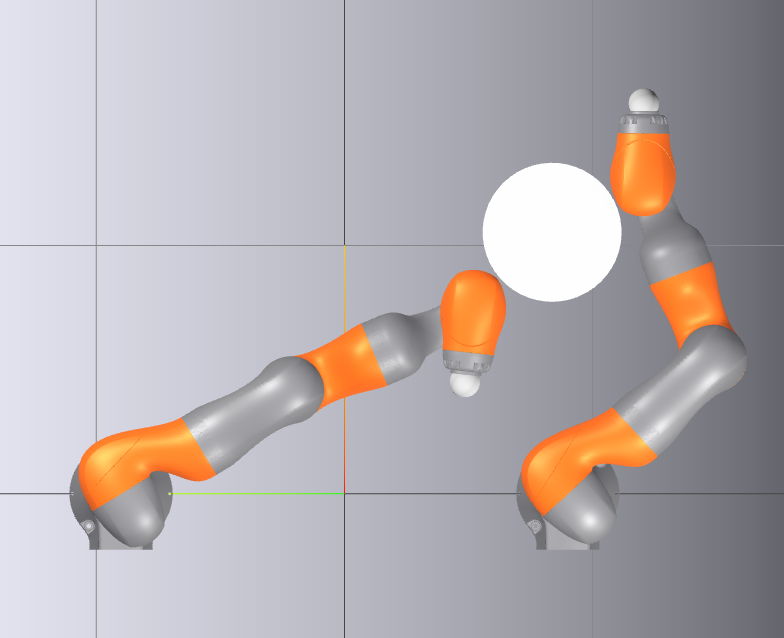}
}
\caption{Illustration of instances where transit fails because manipulator cannot move from pre-transit grasp.}
\label{fig:transit_failure}
\vskip -0.1 true in
\end{figure}

\subsection{GRS Hyperparameters}\label{sec:appendix:hyperparam}

\begin{table}[H]
\centering
\caption{GRS Hyperparameters}
\begin{tabular}{|c|c|c|}
\hline
 Parameter name & Description  & Value \\ \hline
$\alpha$  & Offline: $\mathcal{Q}^o$ coverage & 0.98 \\ \hline
 N  & Offline: number of MRS in graph &  25 \\ \hline
 $\Delta$ & Offline: MRS discretization size & 0.20  \\ \hline
 x range & $\mathcal{Q}^o$ limits & $[0.25, 0.80]$ \\ \hline
 y range  & $\mathcal{Q}^o$ limits  & $[-0.55, 0.55]$  \\ \hline
 $\theta$ range  & $\mathcal{Q}^o$ limits & $[-\pi, \pi]$ \\ \hline
 $\epsilon$ & Offline: IRIS-ZO & 0.01  \\ \hline
 $N_p$ & Offline: IRIS-ZO & 50  \\ \hline
 $N_b$ & Offline: IRIS-ZO & 5  \\ \hline
 $N_t$ & Offline: IRIS-ZO & 5  \\ \hline
  $dt$ & CQDC model timestep & 0.01  \\ \hline
\end{tabular}
\label{tab:appendix:hyperparam:GCSR_hyperparam}
\end{table}

\ifCLASSOPTIONcaptionsoff
  \newpage
\fi



%



\bibliographystyle{IEEEtran}
\bibliography{main}

%








\end{document}